%% file: main.tex
\documentclass[final]{cvpr}
\makeatletter
\@namedef{ver@everyshi.sty}{}
\makeatother

\usepackage{times}
\usepackage{epsfig}
\usepackage{graphicx}
\usepackage{amsmath}
\usepackage{amssymb}

\usepackage{comment}

\usepackage{amscd} 
\usepackage{amsfonts} 
\usepackage{subfigure} 
\usepackage{caption} 
\usepackage{mathtools}

\usepackage[utf8x]{inputenc}

\usepackage{placeins}
\usepackage{algorithm}
\usepackage{algorithmic}
\usepackage{array}
\usepackage{overpic}
\usepackage{bm}
\usepackage{wrapfig}

\usepackage{pgfplots}
\usepackage{pgf,tikz}
\usepackage{tikz-cd} 
\usepackage{xfrac}

\definecolor{darkred}{rgb}{0.85,0,0}
\definecolor{darkgreen}{rgb}{0,0.6,0}
\definecolor{darkred}{rgb}{0.6,0,0}
\definecolor{darkblue}{rgb}{0,0.447,0.741}
\definecolor{darkorange}{rgb}{0.837,0.625,0.112}
\newcommand{\darkgreen}[1]{{\color{darkgreen}{#1}}}  
\newcommand{\darkred}[1]{{\color{darkred}{#1}}}

\newcommand{\camready}[1]{{#1}} 
\newcommand{\veryshortarrow}[1][3pt]{\mathrel{%
   \hbox{\rule[\dimexpr\fontdimen22\textfont2-.2pt\relax]{#1}{.4pt}}%
   \mkern-4mu\hbox{\usefont{U}{lasy}{m}{n}\symbol{41}}}}

\usepackage[pagebackref=true,breaklinks=true,colorlinks,bookmarks=false]{hyperref}

\def\cvprPaperID{5581} 
\def\confYear{CVPR 2021}

\begin{document}

\title{Universal Spectral Adversarial Attacks for Deformable Shapes}

\author{Arianna Rampini\\
Sapienza University of Rome\\
{\tt\small rampini@di.uniroma1.it}
\and
Franco Pestarini\\
Sapienza University of Rome\\
{\tt\small pestarini.1855627@studenti.uniroma1.it}
\and
Luca Cosmo\\
Sapienza University of Rome\\
{\tt\small cosmo@di.uniroma1.it}
\and
Simone Melzi\\
Sapienza University of Rome\\
{\tt\small melzi@di.uniroma1.it}
\and
Emanuele Rodolà\\
Sapienza University of Rome\\
{\tt\small rodola@di.uniroma1.it}
}

\maketitle

\begin{abstract}
Machine learning models are known to be vulnerable to adversarial attacks, namely perturbations of the data that lead to wrong predictions despite being imperceptible. However, the existence of ``universal'' attacks (i.e., unique perturbations that transfer across different data points) has only been demonstrated for images to date. Part of the reason lies in the lack of a common  domain, for geometric data such as graphs, meshes, and point clouds, where a universal perturbation can be defined. In this paper, we offer a change in perspective and demonstrate the existence of universal attacks for geometric data (shapes). We introduce a computational procedure that operates entirely in the spectral domain, where the attacks take the form of small perturbations to short eigenvalue sequences; the resulting geometry is then synthesized via shape-from-spectrum recovery. Our attacks are universal, in that they transfer across different shapes, different representations (meshes and point clouds), and generalize to previously unseen data.
\end{abstract}


\input{sections/introduction.tex}
\input{sections/related.tex}
\input{sections/method.tex}
\input{sections/results.tex}
\input{sections/conclusion.tex}

\section*{Acknowledgments}
\vspace{-0.6ex}
{We gratefully acknowledge Luca Moschella for the technical help. This work is supported by the ERC Grant No.~802554 (SPECGEO) and the MIUR under grant ``Dipartimenti di eccellenza 2018-2022''. 
}

{\small
\bibliographystyle{ieee_fullname}
\bibliography{egbib}
}

\clearpage
\newpage

\begin{center}
      \begin{tabular}[!t]{c}
          \centering  
    {\Large \bf Supplementary Materials \par}
      \vspace*{24pt}
      \large
      \lineskip .5em
      \end{tabular}

      \par
      
\end{center}

\input{sections/architectures}
\input{sections/additional_results.tex}

\end{document}

%% file: sections/introduction.tex
\section{Introduction}\label{sec:introduction}
As machine learning methods become more and more pervasive, so their vulnerabilities are becoming more exposed. In recent years, it has been extensively shown that classifiers are susceptible to so-called adversarial attacks, i.e., misclassifications induced by feeding carefully perturbed data ({\em adversarial examples}) into the trained model. Adversarial examples can be crafted for image, graph, point cloud, and mesh data, as demonstrated by a thriving stream of research output across the computer vision, geometry processing, and machine learning communities. 

\begin{figure}[t]
\centering
     \begin{overpic}[trim=0cm 0cm 0cm 0cm,clip,width=\linewidth]{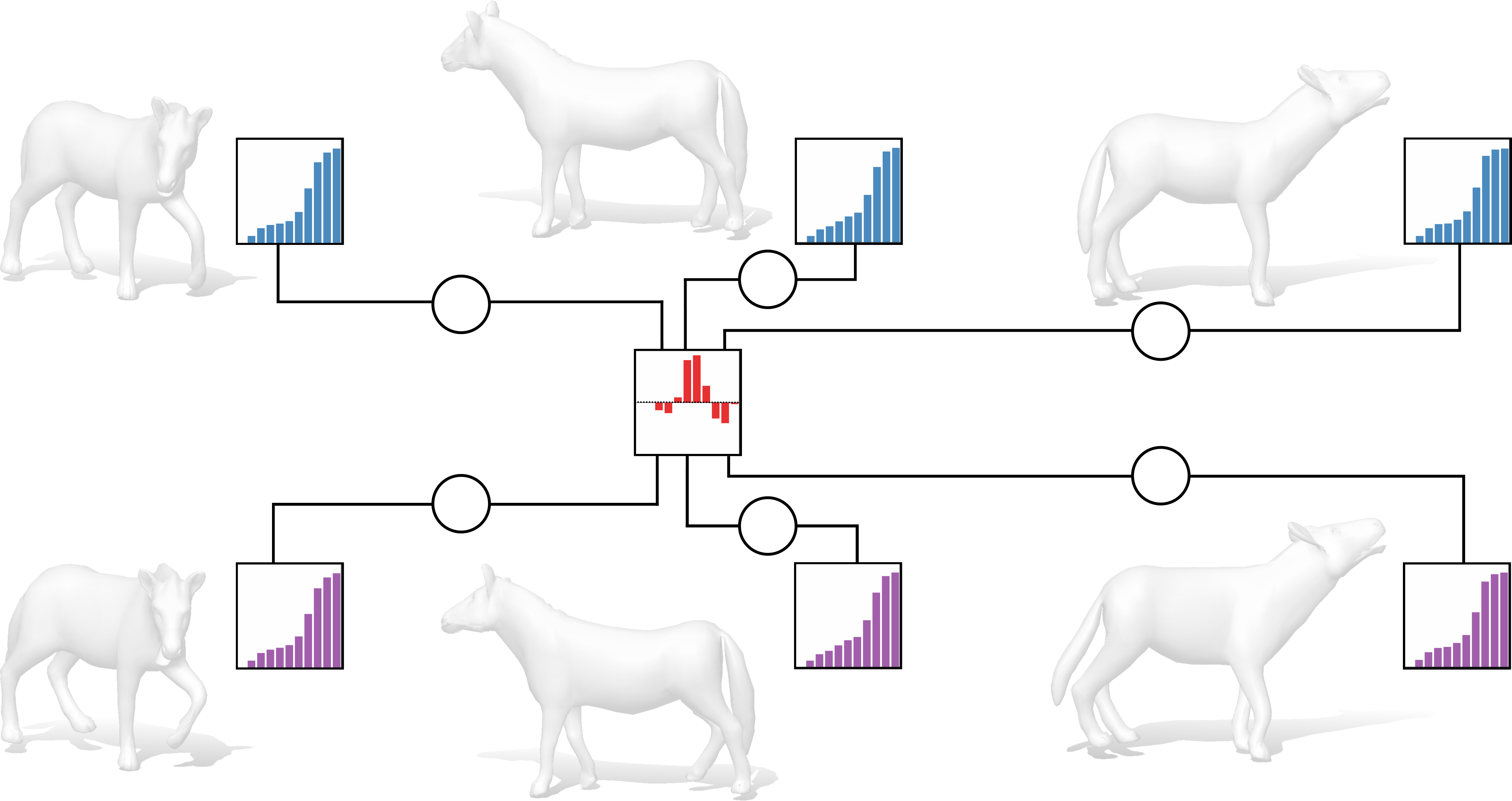}
     \put(3.4,47){\footnotesize{\darkgreen{ horse}}}
    \put(38,49.3){\footnotesize{\darkgreen{ horse}}}
    \put(75,45.8){\footnotesize{\darkgreen{ horse}}}
    \put(4,16.4){\footnotesize{\darkred{dog}}}
    \put(38.5,12.7){\footnotesize{\darkred{cow}}}
    \put(74,16.7){\footnotesize{\darkred{bigcat}}}   
    \put(40.2,25.6){\tiny $0$}   
    \put(33.2,22.5){\tiny $-0.02$}   
    \put(33.3,28.8){\tiny $+0.02$}   
    \put(24,40){$\mathbf{\ldots}$}
    \put(62.2,40){$\mathbf{\ldots}$}
    \put(24,12){$\mathbf{\ldots}$}
    \put(62.2,12){$\mathbf{\ldots}$}
    \put(29.2,31.9){\textbf{+}}
    \put(49.6,33.35){\textbf{+}}
    \put(75.5,29.98){\textbf{+}}
    \put(29.2,18.6){\textbf{=}}
    \put(49.5,17.1){\textbf{=}}
    \put(75.5,20.4){\textbf{=}}
   \end{overpic}
\caption{\label{fig:teaser}Universal spectral attacks on 3D horses from the SMAL dataset (only 3 out of 10 shapes are visualized). {\em Top row}: Original shapes with their first ten Laplacian eigenvalues and the correct labels predicted by a state-of-the-art classifier. The shapes undergo pose deformations, and have different scale, orientation and location in 3D space. {\em Middle row}: A universal perturbation is applied to the eigenvalues. {\em Bottom row}: The resulting shape embeddings synthesized from the perturbed spectra, which are now assigned wrong labels by the classifier.}
\end{figure}
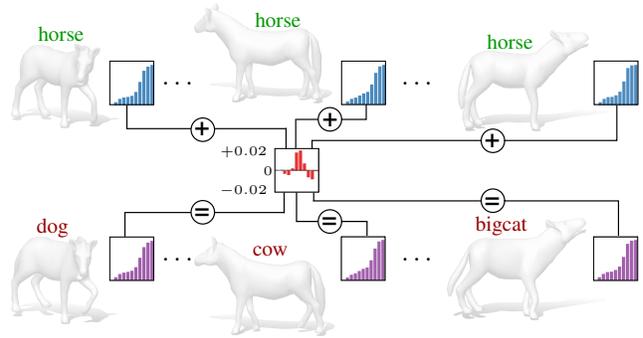

Remarkably, {\em universal} perturbations are also known to exist for image data. For a given classifier $\mathcal{C}$ acting on images of size $w\times h$, a universal perturbation $P\in\mathbb{R}^{w\times h}$ is such that $\mathcal{C}(I+P)\neq\mathcal{C}(I)$ for a large number of images $I$; crucially, $P$ is small under some norm $\|P\|$ so that it is hard to perceive, and is {\em fixed} for all such images. In other words, $P$ is image-agnostic to some extent.
This definition of universal perturbations is made possible by the fact that the operation $I+P$ can be invariably defined for all possible $I$ and $P$, since all $w\times h$ images share the same grid of 2D coordinates.
As soon as one shifts the focus from images to geometric data, the existence of a common space is no longer guaranteed; each individual graph $G$ is a different domain in and by itself, and an operation of the form ``$G+P$'' can only be defined if $G$ and $P$ share the same topology. Therefore, if universality is desired, one has to define a way to transfer the perturbation $P$ across different graphs, while at the same time ensuring it induces misclassification in all cases.

In this paper, we introduce a new paradigm for universal adversarial attacks on geometric data (specifically, meshes and point clouds), in which the perturbation transfer is carried out implicitly. We do so by identifying a common domain as the space of (truncated) Laplacian spectra. This space is compact, since it only consists of short sequences of eigenvalues; it is invariant to isometric deformations (e.g., changes in pose); it only loosely depends on resolution and connectivity of the source geometry; and it is easy and efficient to compute for any given geometric object. Once a {\em universal} perturbation is computed in this space, the individual adversarial examples are recovered via a synthesis process that goes from eigenvalues to 3D coordinates.

%% file: sections/related.tex
\section{Related work}\label{sec:related}
Adversarial attacks were discovered in the seminal paper by Szegedy et al.~\cite{szegedy2013intriguing}, and have since been extensively explored in the image domain~\cite{goodfellow2014explaining,madry2017towards,xie2019feature,sarkar2019enforcing,khoury2019adversarial,zhang2019defending,rakin2018parametric,DBLP:conf/icml/ZhangYJXGJ19,carlini2017towards,athalye2018obfuscated,madry2017towards,rony2019decoupling, papernot2017practical,chen2017zoo,li2019nattack}, natural language processing \cite{gao2018black, chaturvedi2019exploring,jin2019bert}, and reinforcement learning \cite{gleave2019adversarial}, to name just a few.
In this paper, we focus on {\em universal} attacks for {\em geometric} data, hence this section covers relevant prior work addressing the two aspects.

\vspace{1ex}\noindent\textbf{Adversarial attacks on geometric data.}
Compared to the image domain, the literature on adversarial attacks for geometric or topological data is less crowded, but is growing at a steady pace. Attacks on {\em graphs} are relatively more explored due to their relevance in tasks of community detection \cite{chen2017practical}, plausibility and link prediction~\cite{zhang2019data,sun2018data}, and classification \cite{dai2018adversarial,zugner2018adversarial} among others. These attacks operate by modifying the graph topology, i.e., by adding, removing, or rewiring edge connections (see the recent survey \cite{xu2019adversarial} for an in-depth treatment). Our aim is different; instead of attacking the discrete structure representing the 3D shape, for example by changing its triangle connectivity, we seek for attacks that modify the 3D point coordinates. This brings us closer to attacking the underlying surface itself, independently of its specific representation, which in turn endows us with the ability to concoct attacks for both meshes and point clouds within a unified framework.

The literature on adversarial attacks for irregular {\em point cloud} data has also witnessed a recent growth. Works such as \cite{liu2019extending,xiang2019generating,zhang2019defense,hamdi2019advpc} define the adversarial perturbations as small point shifts in 3D space, or as the addition of outlier points to the cloud to confuse the classifier. Since these sparse displacements can lead to noticeable artifacts, additional regularization terms to promote smooth perturbations were introduced in \cite{wen2019geometry,tsairobust}, while in \cite{zhao2020isometry} the perturbation is a global rigid isometry applied to the 3D point cloud.

Works targeting {\em mesh} data are more scarce. In~\cite{xiao2019meshadv}, the authors employ a differentiable renderer to define a perceptual loss, and generate attacks on photorealistic renderings by perturbing the shape texture and geometry. More recently, the work \cite{Mariani20} introduced band-limited perturbations for mesh and point cloud classifiers, resulting in perturbations that are smooth by construction. Since we are also interested in smooth perturbations, we include the smoothness term of \cite{Mariani20} into our construction as well.

None of the aforementioned methods provides a way to seek for universal perturbations, i.e., each perturbation is crafted for a given data sample independently of others, nor can these methods be trivially extended to address the more challenging, universal setting. We will clarify this statement more formally in the sequel.

\vspace{1ex}\noindent\textbf{Universal adversarial attacks} for image classifiers were discovered by Moosavi-Dezfooli et al.~\cite{moosavi2017universal}, who introduced an iterative algorithm to compute universal perturbations over a set of input images. Since then, other approaches have been proposed to find universal perturbations using generative models \cite{hayes2018learning,poursaeed2018generative,reddy2018nag}, more efficient optimization schemes \cite{shafahi2020universal}, based on patches rather than individual pixels~\cite{brown2017adversarial}, or applied to other image-based tasks different from classification \cite{hendrik2017universal,mopuri2018generalizable}; we refer to the recent survey~\cite{chaubey2020universal} for additional examples. On graph structured data, universal attacks were recently considered in \cite{zang2020graph}; however, in their setting, universality is meant across different signals defined on a fixed graph, therefore their attacks do {\em not} transfer among different graphs. To the best of our knowledge, to date, no approaches have been proposed to address universal attacks for graphs or other geometric data such as point clouds and meshes.

For the sake of clarity, we mention here the closely related notion of {\em transferability} of attacks across different architectures, see \cite{liu2016delving,papernot2016transferability} for examples with image-based classifiers, and \cite{hamdi2019advpc} for point clouds. This is different than universality, which is instead meant across data samples (the scope of this paper), rather than across learning models. 

\subsection{Contribution}
Our main contributions can be summarized as follows:
\begin{itemize}
    \item We demonstrate, for the first time, the existence of universal adversarial perturbations for non-rigid 3D geometric data;
    \item We introduce a computational procedure for finding such perturbations, which operates in the spectral domain, and follows an analysis--synthesis paradigm;
    \item We show that our attacks are universal in two ways: (i) across different shapes, and (ii) across different representations, such as meshes and point clouds;
    \item We show the generalization property of our attacks to previously unseen data.
\end{itemize}

%% file: sections/method.tex
\section{Universal spectral perturbations}\label{sec:method}
Following prior work on universal attacks, our framework assumes white-box access to a given classifier, since we backpropagate the error through its parameters (which are held fixed throughout the entire optimization).
Further, we focus on {\em untargeted} attacks; namely, we do not specify a target class for the misclassification, but only require the classifier to change its prediction.

\subsection{Problem setting \& motivation}
Given a pre-trained classifier $\mathcal{C}$ and a set of objects $\{X_i\}$, our objective is to find a perturbation $P$ such that:
\begin{enumerate}
    \item $\mathcal{C}(X_i + P)\neq\mathcal{C}(X_i)$ for most $i$ and for a proper definition of the `$+$' operation (\textbf{universality});\label{obj1} 
    \item $P$ is small in some sense, since it must remain unnoticed (\textbf{noticeability}).\label{obj2}
\end{enumerate}
The goals set above generalize those found in \cite{moosavi2017universal} to a broader setting.
If the objects $\{X_i\}$ are plain images of fixed size as in \cite{moosavi2017universal}, then the sum operation is well defined pixel-wise, since both $P$ and the image set $\{X_i\}$ belong to the same vector space. However, if each $X_i$ is an instance of non-flat structured data, one faces a number of issues.

\begin{figure}[t]
\centering
     \begin{overpic}[trim=0cm 0cm 0cm 0cm,clip,width=0.80\linewidth]{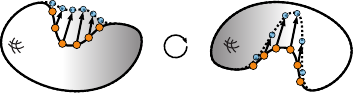}
   \end{overpic}
\caption{\label{fig:extrinsic}An extrinsic perturbation can not be universal and transformation-invariant at the same time. Even if a map is available for each pair of shapes, using it to estimate the right transformation for the perturbation is possible in the rigid case, but much harder in the general non-rigid case, and not guaranteed to lead to misclassification in both cases.}
\end{figure}

Assume for simplicity that each ${X_i\in\mathbb{R}^{n\times 3}}$ is a 3D point cloud with $n$ points. Following previous adversarial schemes for point clouds \cite{liu2019extending,xiang2019generating,zhang2019defense,hamdi2019advpc,wen2019geometry,tsairobust}, a perturbation $P\in\mathbb{R}^{n\times 3}$ can be defined as a displacement field such that $X_i+P$ is a slight modification of the point positions of $X_i$ in 3D space. However, such a $P$ can not be optimized to be universal, since it can not be directly added to a different shape $X_j$ with $j\neq i$. 
\camready{ First, the sum $X_j+P$ only makes sense if $X_i$ and $X_j$ have the same point ordering, or equivalently, if a dense point-to-point map is available between them. Second, even if a map is available, this type of attack can not be deformation-invariant: since a per-vertex perturbation is extrinsic by definition, it depends on the specific 3D coordinates to which it is applied (see Figure~\ref{fig:extrinsic}). Therefore, a successful per-vertex attack on mesh $X_i$ will not remain successful if $X_i$ is rotated or isometrically deformed, even if these transformations preserve the mesh topology. }


%
%

\subsection{Our approach}
The main issue of the aforementioned approach is that it models perturbations as {\em extrinsic} quantities, i.e., which depend on the specific way in which the 3D objects are embedded into the ambient Euclidean space.

To address this issue, we propose to shift to an {\em intrinsic} representation. Let us be given a set of shapes $\mathcal{S}=\{X_i\}$. For each shape $X_i\in\mathcal{S}$ we define its {\em spectral representation} of length $k$ as the sequence:
\begin{align}
    \sigma(X_i) = (\lambda_1^i, \lambda_2^i, \dots ,\lambda^i_k)\,,
\end{align}
where the $\lambda$'s are the first $k$ eigenvalues of the Laplace-Beltrami operator of $X_i$, ordered increasingly. The Laplacian eigenvalues capture geometric information of the shape and can be computed easily; importantly, they do {\em not} depend on how the shape is embedded in 3D, but they are an intrinsic quantity that is invariant to non-rigid isometries. 

\paragraph*{Objective \ref{obj1}: Universality.}
We define our universal perturbation $\rho$ to be a local solution to the following nonlinear optimization problem:
\begin{align}
    \label{eq:general_formula}
    \min_{\substack{\rho \in \mathbb{R}^{k}\\\mathcal{P}_i}} & ~~ \sum_{X_i\in\mathcal{S}} \| \sigma(X_i)(1+\rho) - \sigma(\mathcal{P}_{i}(X_i)) \|^{2}_{2}\\
    \label{eq:constraint}
    \mathrm{s.t.}& ~~ \mathcal{C}(\mathcal{P}_{i}(X_i)) \neq \mathcal{C}(X_i) \quad \forall X_i\in\mathcal{S} 
\end{align}
%

Note that $\rho\in\mathbb{R}^k$ is universal and is an element-wise multiplicative perturbation in the {\em spectral} domain, while $\mathcal{P}_{i}(X_{i})$ are shape-specific extrinsic perturbations for each shape. The constraints of Eq.~\eqref{eq:constraint} ensure that all the shapes in the optimization are misclassified.
The spectral perturbation is multiplicative, rather than additive, so that it does not depend on the absolute scale of the eigenvalues.

To get a better grasp of what the energy of Eq.~\eqref{eq:general_formula} enforces, we illustrate its action via the following commutative diagram:
\[\begin{tikzcd}
X_i \arrow{r}{\sigma} \arrow[swap]{d}{\color{darkblue}{\mathcal{P}_i}} & (\lambda^i) \arrow{d}{{\color{darkorange}{\rho}}} \\
\tilde{X}_i \arrow{r}{\sigma} & (\tilde{\lambda}^i)
\end{tikzcd}
\]
The diagram expresses the fact that perturbing the eigenvalues of a given shape $X_i$ (upper path) is equivalent to first perturbing the shape embedding itself, and then computing its eigenvalues (lower path). 
This is known to be true for small perturbations; a classical result of Bando and Urakawa~\cite{bando1983generic} states that Laplacian eigenvalues change continuously with the surface metric, meaning that a small perturbation of the spectrum corresponds to a small perturbation of the geometry. 

Both the spectral perturbation $\color{darkorange}\rho$ and the spatial perturbations ${\color{darkblue}\mathcal{P}_i}(X_i)$ are unknown and must be solved for; however, the former is \textcolor{darkorange}{shape-agnostic} and fixed for all $i$, while the latter is \textcolor{darkblue}{shape-dependent}.
We seek for the set of extrinsic modifications to the geometries in $\mathcal{S}$, that simultaneously give rise to the {\em same} change in the eigenvalues.

\vspace{1ex}\noindent\textbf{\textit{Remark.}}
The optimal $\color{darkorange}\rho^\ast$ minimizing Eq.~\eqref{eq:general_formula}-\eqref{eq:constraint} is a well-defined {\em universal} perturbation, since it applies to all shapes in the optimization set $\mathcal{S}$. In particular, one can discard the shape-dependent $\mathcal{P}_i$'s, and verify misclassification for all $X_i\in\mathcal{S}$ by decoding $\sigma(X_i)(1+{\color{darkorange}\rho^\ast})$ to a 3D shape (we give an algorithm in Sec.~\ref{sec:alg}).

\vspace{2ex}
The remark above supports our main claim on the existence of universal adversarial perturbations for non-rigid 3D geometric data.
Furthermore, as we empirically show in our experiments, the universal spectral perturbations $\color{darkorange}\rho^\ast$ also exhibit {\em generalization} outside of the optimization set in several cases.

\paragraph*{Objective \ref{obj2}: Noticeability.}
We model the {\em per-shape} perturbation $\mathcal{P}_{i}(X_i)$ as an extrinsic displacement $X_i+P_i$. Adversarial attacks on images explicitly impose an upper bound $\|P_i\|<\epsilon$ (see, e.g.,~\cite{moosavi2017universal}) to ensure imperceptible perturbations. Here we appeal instead to the theoretical result, also mentioned previously, that small changes in the geometry correspond to small changes in the eigenvalues \cite{bando1983generic}. This expectation is encoded in our energy of Eq.~\eqref{eq:general_formula}. Therefore, by minimizing this energy, we are also implicitly bounding the perturbation strength.

Further, we follow the smoothness principle of~\cite{Mariani20,tsairobust,wen2019geometry}, which aims to ensure that each $P_i$ is as smooth as possible. This corresponds to imposing a bound on the gradient norm $\|\nabla P_i\|$, preventing jittered perturbations.
In particular, we adopt subspace parametrization \cite{Mariani20} due to its simplicity. Each $P_i$ is expressed as a linear combination of smooth vector fields:
\begin{align}
    \label{eq:3Ddisplacement}
    \mathcal{P}_{i}(X_i) = X_i + \Phi_i {\alpha}_{i}\,,
\end{align}
where $\Phi_i$ is a $n\times b$ matrix whose columns are the first $b$ Laplacian eigenfunctions of $X_i$ (with $b \ll n$, where $n$ is the total number of vertices), and ${\alpha}_i$ is a $b\times 3$ matrix of expansion coefficients. For smaller values of $b$, one gets a smoother deformation field.
%
%
%
%
%
%
This band-limited representation of the displacement only requires solving for  $3b\ll 3n$ coefficients per shape; furthermore, it ensures smoothness (bounded gradient) as Laplacian eigenfunctions are optimal for representing smooth functions (see \cite[Th.~3.1]{aflalo2015optimality}).


The complete optimization problem reads:
\begin{align}
    \label{eq:general_formula2}
    \min_{\substack{\rho \in \mathbb{R}^{k}\\\{\alpha_i \}_{i}}} &~~ \sum_{X_i\in\mathcal{S}} \| \sigma(X_i)(1+\rho) - \sigma(X_i + {\Phi}_i{\alpha}_{i}) \|^{2}_{2}\\
    \label{eq:constraint2}
    \mathrm{s.t.}& ~~ \mathcal{C}({X}_i + {\Phi}_i{\alpha}_{i}) \neq \mathcal{C}(X_i) \quad \forall X_i\in\mathcal{S}
\end{align}
which involves in total $k$ optimization variables for the spectral perturbation $\rho$, and $3b\cdot |\mathcal{S}|$ variables for the spatial perturbation coefficients $\alpha_i$; in our tests, we typically use $b=20$ and $k=3b$. These numbers do {\em not} depend on the number of points of the shapes $X_i$, hence we can afford optimizing over shapes with varying resolutions.

\subsection{Properties of spectral perturbations}
Before moving on to the algorithmic details, we list here a few important properties of our formulation.
The key idea behind this approach lies in the realization that the space of eigenvalues can serve as a convenient common domain, where different geometric data can be easily represented. The spectral domain carries important invariances that are directly inherited by our perturbations:
\begin{itemize}
    \item We do not need an input correspondence between the shapes, nor do we have to solve for one. This also goes beyond adversarial perturbation methods for images, where one exploits the correspondence given ``for free'' by the canonical ordering of the pixel grid;
    \item Since Laplacian eigenvalues are robust against varying point density and resolution, 
    our optimization does not require the shapes to have the same number of points or same resolution;
    \item Since Laplacian eigenvalues can be computed both for meshes and point clouds, spectral perturbations do not require a special treatment depending on the geometry representation.
\end{itemize}

Laplacian spectra are isometry-invariant, hence their perturbation is expected to have similar effects on isometric shapes. Similarly, since we use multiplicative perturbations, we are also invariant to scale changes of the eigenvalues, and in turn, to scale changes of the 3D shapes.
We empirically confirm these properties in the experimental section.


\subsection{Algorithm}\label{sec:alg}
We follow the general approach of Carlini and Wagner~\cite{carlini2017towards} to minimize problem~\eqref{eq:general_formula2}, and pass to the unconstrained minimization:
\begin{align}
    \label{eq:general_formula3}
    \hspace{-0.34cm}
    \min_{\substack{\rho \in \mathbb{R}^{k}\\\{\alpha_i \}_{i}}}  \hspace{-0.05cm}
    \sum_{X_i\in\mathcal{S}}\hspace{-0.08cm} \| \sigma(X_i)(1\hspace{-0.07cm}+\hspace{-0.07cm}\rho) \hspace{-0.05cm}- \hspace{-0.05cm}\sigma(X_i\hspace{-0.07cm} + \hspace{-0.07cm}{\Phi}_i{\alpha}_{i}) \|^{2}_{2} \hspace{-0.05cm} +\hspace{-0.05cm} c \mathcal{A}(X_i,\alpha_i)
\end{align}
where $\mathcal{A}$ is an adversarial penalty relaxing the constraints of Eq.~\eqref{eq:constraint2}, and defined as follows:
\begin{align}
    \label{eq:adversarial_loss}
    \mathcal{A}(X_i&,\alpha_i) = \\
    \mu&(Z(X_i,\alpha_i)_{\mathcal{C}(X_i)} \hspace{-0.05cm} - \hspace{-0.05cm} \max\{Z(X_i,\alpha_i)_j\hspace{-0.05cm} :\hspace{-0.05cm} j\neq \mathcal{C}(X_i)\}) \nonumber
\end{align}


Here $Z(X_i,\alpha_i)$ is the unnormalized log-probability vector predicted by classifier $\mathcal{C}$ for the shape $({X}_i + {\Phi}_i{\alpha}_{i})$, and $\mu(x) = \max(x,-m)$ is a function sending the penalty to zero once a given misclassification margin $m$ is hit.
The contribution of the adversarial penalty to the minimization problem is weighted by the trade-off parameter $c$.

Solving this unconstrained problem does not guarantee that all the shapes are misclassified and, in general, such a perturbation is not guaranteed to always exist. Nevertheless, in practice the optimized perturbation leads to misclassification for most of the shapes, as we show in our experiments.

\paragraph*{Optimization.}
For each shape $X_i$, we discretize its Laplace-Beltrami operator as a positive semi-definite matrix using the classical cotangent scheme~\cite{pinkall1993computing}, whose eigenvalues and eigenfunctions can be computed with standard sparse eigensolvers. \camready{The optimization variables of Eq.~\eqref{eq:general_formula3} are optimized for with the Adam optimizer \cite{kingma2014adam}, which is robust to local minima. This involves computing the quantities $\sigma(X_i+\Phi_i\alpha_i)$ at each iteration, i.e., the eigenvalues of the deformed shapes, as well as their derivatives with respect to the deformation coefficients $\alpha_i$.} For the eigenvalue derivatives, we use the closed form expressions of Magnus~\cite{magnus1985differentiating}.
%
%
%
%
Each iteration takes approximately 1s on an i7 9700k CPU (dominated by eigenvalue decomposition); for an average number of 500 iterations per optimization, the average runtime to find a universal perturbation on a set of 15 shapes is $\sim$1h.

\vspace{1ex}\noindent\textbf{Generalization to new samples.}
Once a spectral perturbation $\rho$ is estimated for a small set of shapes, it can be applied to new shapes and still fool the classifier. This was also observed for the image-based universal attacks of~\cite{moosavi2017universal}. However, in the image domain, applying a universal perturbation to a new data sample is a simple addition of two images. In our case, the perturbation is not additive in the spatial domain, but multiplicative in the spectral domain. 

Given $\rho$ and a new shape $Y$, we follow the paradigm:
\begin{align}Y \mapsto (\lambda_i)_{i=1}^k \mapsto (\lambda_i + \lambda_i\rho_i)_{i=1}^k \mapsto \tilde{Y}\,,
\end{align}
The first two steps are straightforward and can be easily computed. The last step requires resynthesizing the geometry from the perturbed spectrum -- an inverse problem known in mathematical physics as `hearing the shape of the drum'~\cite{kac}, and recently tackled by `isospectralization' techniques in~\cite{isosp,instant2020}.

\begin{figure}
    \centering
     \begin{overpic}[trim=0cm 0cm 0cm 0cm,clip,width=\linewidth]{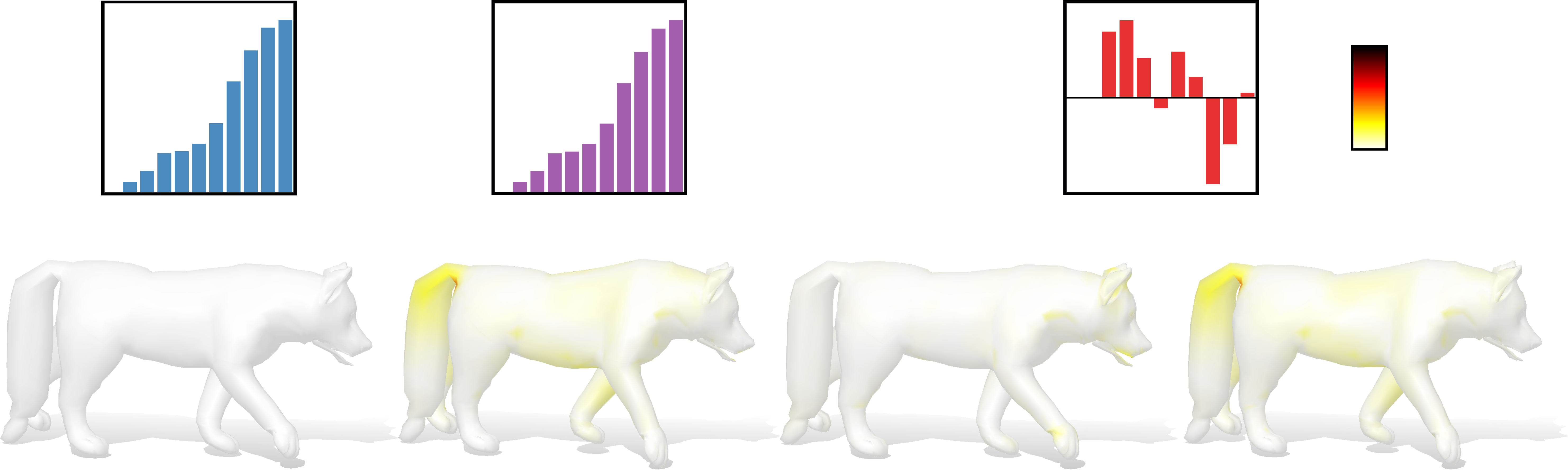}
     \put(6,-2.5){\footnotesize{Original}}
     \put(29.5,-2.5){\footnotesize{Perturbed}}
     \put(58,-2.5){\footnotesize{\cite{isosp}}}
     \put(82,-2.5){\footnotesize{\textbf{Ours}}}
    \put(12,31){\footnotesize{$\lambda$}}
    \put(32.5,31){\footnotesize{$\lambda + \rho \lambda$}}
    \put(73,31){\footnotesize{$\rho$}}
    \put(65.5,23){\tiny $0$}   
    \put(58.5,18){\tiny $-0.02$}   
    \put(58.8,28){\tiny $+0.02$}
    \put(89.5,20){\tiny $0$}   
    \put(89.5,26){\tiny $0.2$} 
    \end{overpic}
    \caption{For a source shape ${X\in\mathcal{S}}$ with spectrum $\lambda$ (left), we compute a perturbation $\rho$ by optimizing Eq.~\eqref{eq:general_formula3} over $\mathcal{S}$, and obtain the Perturbed shape (second from the left, observe the foreleg movement). If we now discard the Perturbed embedding and try to recover it using \cite{isosp} with $\lambda+\rho\lambda$ as a target, we get a wrong solution (third from the left) which aligns correctly the eigenvalues, but fails to recover the correct deformation. Our approach based on optimizing for a smooth deformation recovers the Perturbed shape almost exactly (rightmost).
    %
    }
    \label{fig:isospec}
\end{figure}
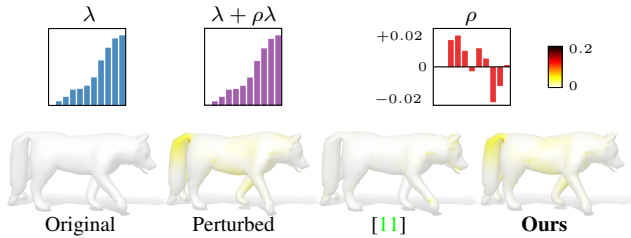

To address this, we simply run the optimization of Eq.~\eqref{eq:general_formula3} without the adversarial term and with fixed $\rho$, which is now given. This way, we optimize only for the coefficients $\alpha$, which define a smooth transformation for the geometry of $Y$. Optimizing over smooth geometric perturbations has a regularization effect as it greatly reduces the space of possible embeddings, making this `isospectralization' problem easier to solve than in~\cite{isosp,instant2020}; in these works, a solution is sought from scratch over all possible point configurations in 3D, making the optimization much harder and prone to poor reconstructions, as shown with a comparison in Figure~\ref{fig:isospec}.
%

%% file: sections/results.tex
\section{Experimental evaluation}\label{sec:results}
In this section we report quantitative results and show qualitative examples of universal spectral perturbations, demonstrating their efficacy and empirically confirming their main properties.

\vspace{1ex}\noindent\textbf{Datasets.}
We tested with two recent and extensive datasets of non-rigid 3D shapes: the \textbf{SMAL} dataset of 3D animals~\cite{SMAL}, and the \textbf{CoMA} dataset of human face expressions~\cite{COMA}.
The former is composed by 600 meshes of 5 animal species in different poses, generated via a parametric model. For fair comparisons, we used the same shapes and experimental setup proposed in~\cite{Mariani20}, using 480 shapes for training the classifiers, and the remaining 120 for test. The classification task assigns each shape to a specific animal category. CoMA is a 4D dataset containing sequences of 3D shapes of 13 different people performing 13 different facial expressions. We used the same train/test split proposed in~\cite{ranjan2018generating} to train the classifiers, where the task is to classify on subject identity.

\vspace{1ex}\noindent\textbf{Classifiers.}
We perform our attacks on two different state-of-the-art classifiers for 3D shapes:
\begin{enumerate}
    \item A convolutional mesh classifier with the  architecture of~\cite{COMA}, where the convolution is based on fast Chebyshev filters~\cite{defferrard2016convolutional}. This learning model is powerful, but needs consistent meshing and correspondence at training time. We refer to this classifier as \textbf{ChebyNet};
    \item A PointNet based classifier~\cite{qi2017pointnet}. This architecture is more general, as it is able to handle unorganized point clouds, possibly with different numbers of points. We refer to this classifier as \textbf{PointNet}.
\end{enumerate}
Both classifiers are trained to classify the subject identity for CoMA data, and the animal species for SMAL data, irrespective of pose.
During the training phase, we augmented each dataset by randomly rotating and translating the shapes, and jittering the vertex positions.

%
\begin{figure}
    \centering
     \begin{overpic}[trim=0cm 0cm 0cm -0.5cm,clip,width=0.92\linewidth]{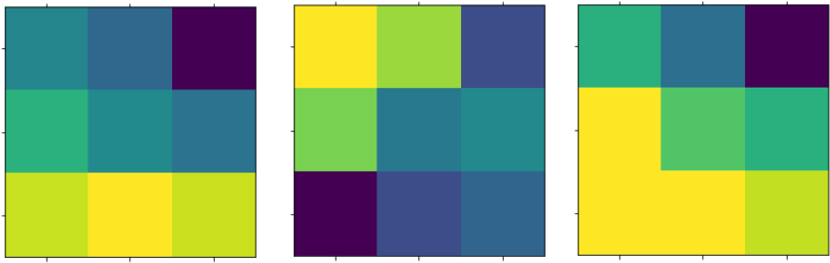}
     \put(3.5,33){\footnotesize \textbf{curv. distortion}}
     \put(2.5,24.5){\footnotesize \color{white}{\textbf{1.31}}}
     \put(3,14.5){\footnotesize \color{white}{\textbf{1.6}}}
     \put(2.5,4.5){\footnotesize \color{white}{\textbf{2.03}}}
     \put(12.5,24.5){\footnotesize \color{white}{\textbf{1.12}}}
     \put(12.5,14.5){\footnotesize \color{white}{\textbf{1.35}}}
     \put(12.5,4.5){\footnotesize \color{black}{\textbf{2.16}}}
     \put(22.5,24.5){\footnotesize \color{white}{\textbf{0.61}}}
     \put(23.5,14.5){\footnotesize \color{white}{\textbf{1.2}}}
     \put(22.5,4.5){\footnotesize \color{white}{\textbf{2.04}}}
     \put(42.5,33){\footnotesize \textbf{$L^2$ norm}}
     \put(36,24.5){\footnotesize \color{black}{\textbf{0.087}}}
     \put(36,14.5){\footnotesize \color{white}{\textbf{0.082}}}
     \put(36,4.5){\footnotesize \color{white}{\textbf{0.065}}}
     \put(46.2,24.5){\footnotesize \color{white}{\textbf{0.083}}}
     \put(46.2,14.5){\footnotesize \color{white}{\textbf{0.074}}}
     \put(47.2,4.5){\footnotesize \color{white}{\textbf{0.07}}}
     \put(57.5,24.5){\footnotesize \color{white}{\textbf{0.07}}}
     \put(56.5,14.5){\footnotesize \color{white}{\textbf{0.075}}}
     \put(56.5,4.5){\footnotesize \color{white}{\textbf{0.072}}}
     \put(75.5,33){\footnotesize \textbf{success rate}}
     \put(71.5,24.5){\footnotesize \color{white}{\textbf{91.7}}}
     \put(71.5,14.5){\footnotesize \color{black}{\textbf{98.3}}}
     \put(71.5,4.5){\footnotesize \color{black}{\textbf{98.3}}}
     \put(81.5,24.5){\footnotesize \color{white}{\textbf{86.7}}}
     \put(81.5,14.5){\footnotesize \color{white}{\textbf{93.3}}}
     \put(81.5,4.5){\footnotesize \color{black}{\textbf{98.3}}}
     \put(92.7,24.5){\footnotesize \color{white}{\textbf{80}}}
     \put(91.5,14.5){\footnotesize \color{white}{\textbf{91.7}}}
     \put(91.5,4.5){\footnotesize \color{white}{\textbf{96.7}}}
     \put(-4.5,24.5){\tiny $15$}
     \put(-4.5,14.5){\tiny $20$}
     \put(-4.5,4.5){\tiny $30$}
     %
     %
     \put(3.5,-3){\tiny $45$}
     \put(13.5,-3){\tiny $60$}
     \put(23.5,-3){\tiny $90$}
     \put(38,-3){\tiny $45$}
     \put(48,-3){\tiny $60$}
     \put(58,-3){\tiny $90$}
     \put(72.5,-3){\tiny $45$}
     \put(82.5,-3){\tiny $60$}
     \put(92.5,-3){\tiny $90$}
     \put(-2.5,-2.6){\sfrac{$b$}{$k$}}
   \end{overpic}
    \caption{Sensitivity to parameters (SMAL dataset, PointNet classifier). We evaluate each quantitative measure at increasing values of $b=(15, 20, 30)$ and number of eigenvalues $k=(45, 60, 90)$. Large numbers for curvature distortion imply more noticeable perturbations. From the success rate we observe a trend: the spectral bandwidth $k$ should not be too large, and the deformation not too smooth. However, lack of smoothness also leads to larger noticeability.}
    \label{fig:k_and_b}
\end{figure}

\begin{figure}[b]
\centering
     \begin{overpic}[trim=0cm 0cm 0cm 0cm,clip,width=\linewidth]{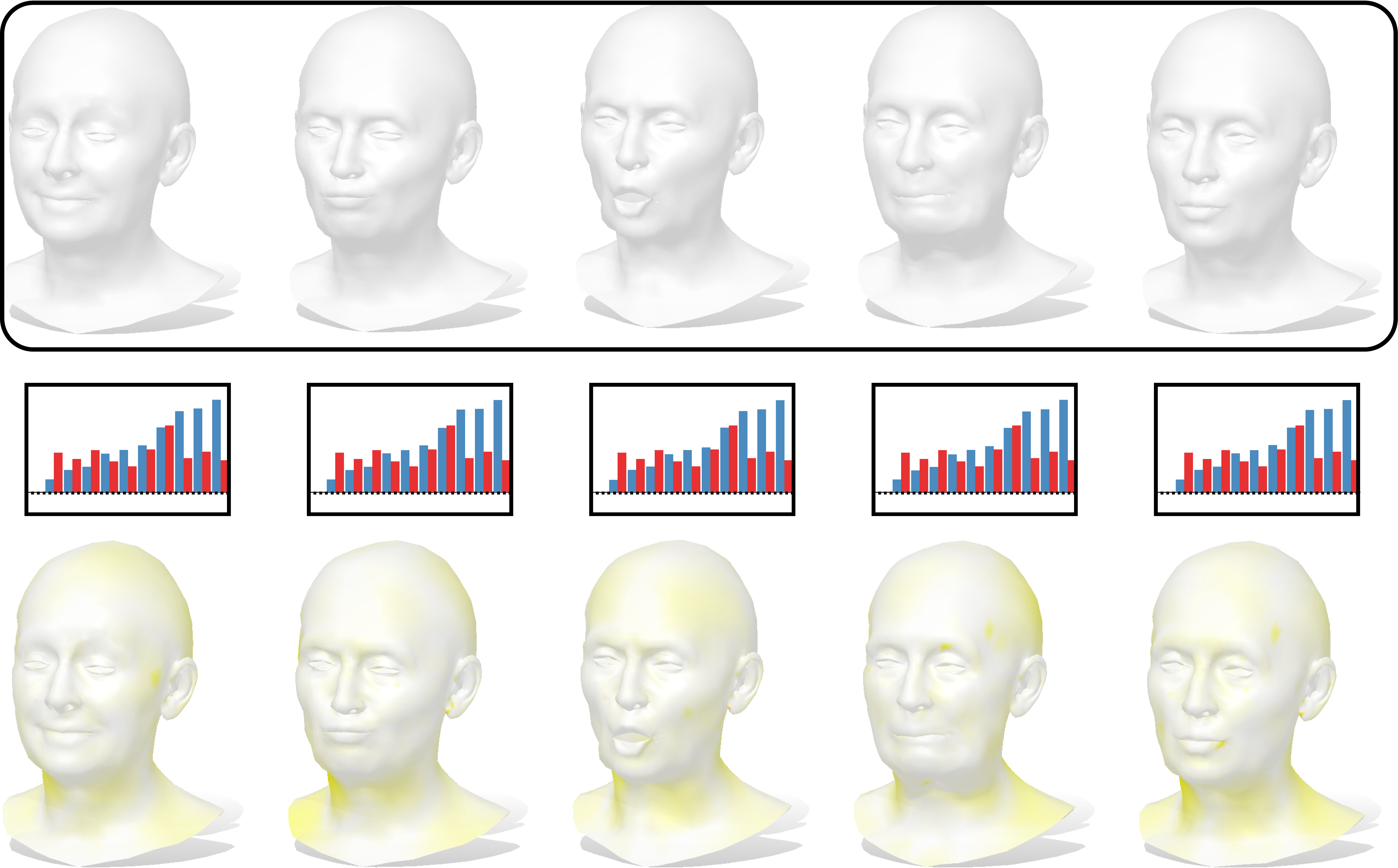}
     \put(46,63){\footnotesize{\darkgreen{ ID 4}}}
     
     \put(4,-3){\footnotesize{\darkred{ ID 2}}}
     \put(24.5,-3){\footnotesize{\darkred{ ID 2}}}
     \put(45,-3){\footnotesize{\darkred{ ID 2}}}
     \put(65.5,-3){\footnotesize{\darkred{ ID 2}}}
     \put(86,-3){\footnotesize{\darkred{ ID 2}}}
     
    \put(0,26.5){\tiny $0$}   
    \put(-3,33){\tiny $720$} 
   \end{overpic}
\caption{\label{fig:COMAqualitative} Examples of a universal adversarial attack on ChebyNet from a class of the CoMA dataset. Top to bottom: original shapes, barplot with their spectra in blue and their perturbation $\rho$ in red ($\rho$ is scaled by a factor $10^{4}$ for visualization purposes), deformed shapes.}
\end{figure}
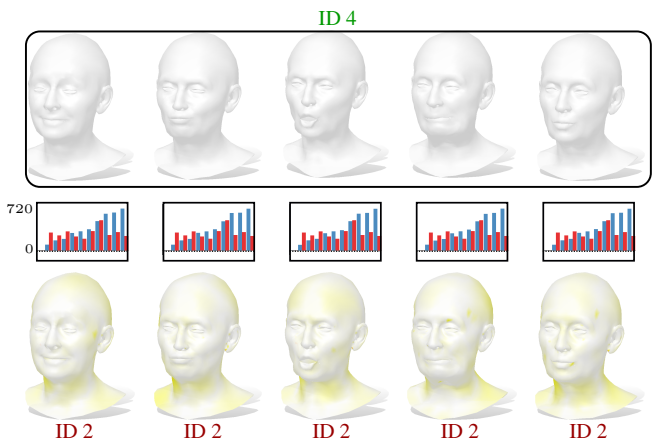

\subsection{Sensitivity to parameters}
We expose two parameters: the spectral bandwidth $k$, that is the number of eigenvalues that undergo the spectral perturbation, and the spatial bandwidth $b$, that is the number of eigenfunctions to represent the spatial perturbation.
Both affect the noticeability and success rate of the attack.

A small value for $b$ leads to smoother and less noticeable perturbations, but makes it harder to find universal ones. On the other hand, large values allow for stronger deformations, but which are also more universal. This is typical of universal perturbations, which have been observed to be more noticeable than per-instance perturbations also in the image domain~\cite{moosavi2017universal}.
Parameter $k$ expresses the degree of universality that we require from the attack, since each of the $k$ dimensions of $\rho$ encodes a constraint for the perturbation. A small $k$ leads to more global deformations, leaving the attack free to apply local shape-dependent corrections. Increasing $k$ makes it harder to obtain a successful universal attack, since a longer perturbation vector $\rho$ imposes more geometric constraints on the attack.


We performed a systematic study of $b$ and $k$, quantifying noticeability through two deformation measures: \emph{curvature distortion}, defined as the average absolute difference between the mean curvature at corresponding vertices in the original and perturbed shape; and \mbox{$L^2$-norm}, defined as the average Euclidean distance between corresponding vertices in the original and perturbed shape. The {\em success rate} is the percentage of attacks that give rise to misclassification.

Quantitative results are reported in Figure~\ref{fig:k_and_b}, revealing a trade-off between the amount of curvature distortion and the success rate. On the contrary, the $L^2$-norm decreases with $b$; this is consistent with what was shown in~\cite{Mariani20}, since smoother deformations force the attack to move a bigger proportion of the shape, as the classifier can not be fooled with small local perturbations. Based on this, we claim that $L^2$-norm is probably not a good metric for capturing noticeability on deformable shapes, since localized deformations are usually more disturbing to the human observer. 
For all our experiments, we use $b=20$ and $k=60$ as a good trade-off between noticeability and success rate; in our plots, we only show $k=10$ eigenvalues for visualization purposes.

\begin{table}[]
        \caption{Comparison between our method and non-universal approaches. We observe that a slight drop in accuracy and an increase of the deformation strength is needed, in order to gain a universal encoding of the attacks.}
    \label{tab:comparison}
    
    \renewcommand{\arraystretch}{0.8}
    \centering
    \begin{tabular}{l|c c c}
    
    \hline
    \rule{0pt}{0.8em} \footnotesize \textbf{SMAL}         & \footnotesize success rate & \footnotesize curv. dist. &  \footnotesize $L^2$-norm \\
    \hline
    \footnotesize\textsc{shape-dependent}\\
    \footnotesize \cite{Mariani20} (ChebyNet)  & \footnotesize 100\% &\footnotesize  2.51 & \footnotesize 3.6e-2 \\
    \footnotesize ChebyNet                    & \footnotesize 100\% & \footnotesize 2.91 & \footnotesize 7.7e-2 \\
    \hline                    
   \footnotesize  \textsc{universal}\\
   \footnotesize  ChebyNet                    &   \footnotesize 100\% &  \footnotesize 2.49 & \footnotesize 9.0e-2\\
   \footnotesize  PointNet                    &  \footnotesize   93.3\% & \footnotesize  1.31 & \footnotesize 5.9e-2\\
    \hline
    \hline
    \rule{0pt}{0.8em} \footnotesize \textbf{CoMA}                        & \footnotesize success rate & \footnotesize curv. dist. & \footnotesize $L^2$-norm\\
    \hline
    \footnotesize \textsc{shape-dependent}\\
   \footnotesize  \cite{Mariani20} (ChebyNet)  & \footnotesize 94.3\% & \footnotesize 3.30  & \footnotesize 8.5e-3\\
   \footnotesize  ChebyNet                    & \footnotesize 91.7\% & \footnotesize 1.61  & \footnotesize 1.7e-3\\
    \hline                    
   \footnotesize  \textsc{universal}\\
   \footnotesize  ChebyNet                    &  \footnotesize  91.7\% & \footnotesize  2.30 & \footnotesize 2.8e-3\\
   \footnotesize  PointNet                    &   \footnotesize  100\% &  \footnotesize 5.71 & \footnotesize 6.9e-3\\
    \hline
    \end{tabular}

\end{table}

\subsection{Universality}
\begin{figure}[t]
\centering
     \begin{overpic}[trim=0cm 0cm 0cm 0cm,clip,width=\linewidth]{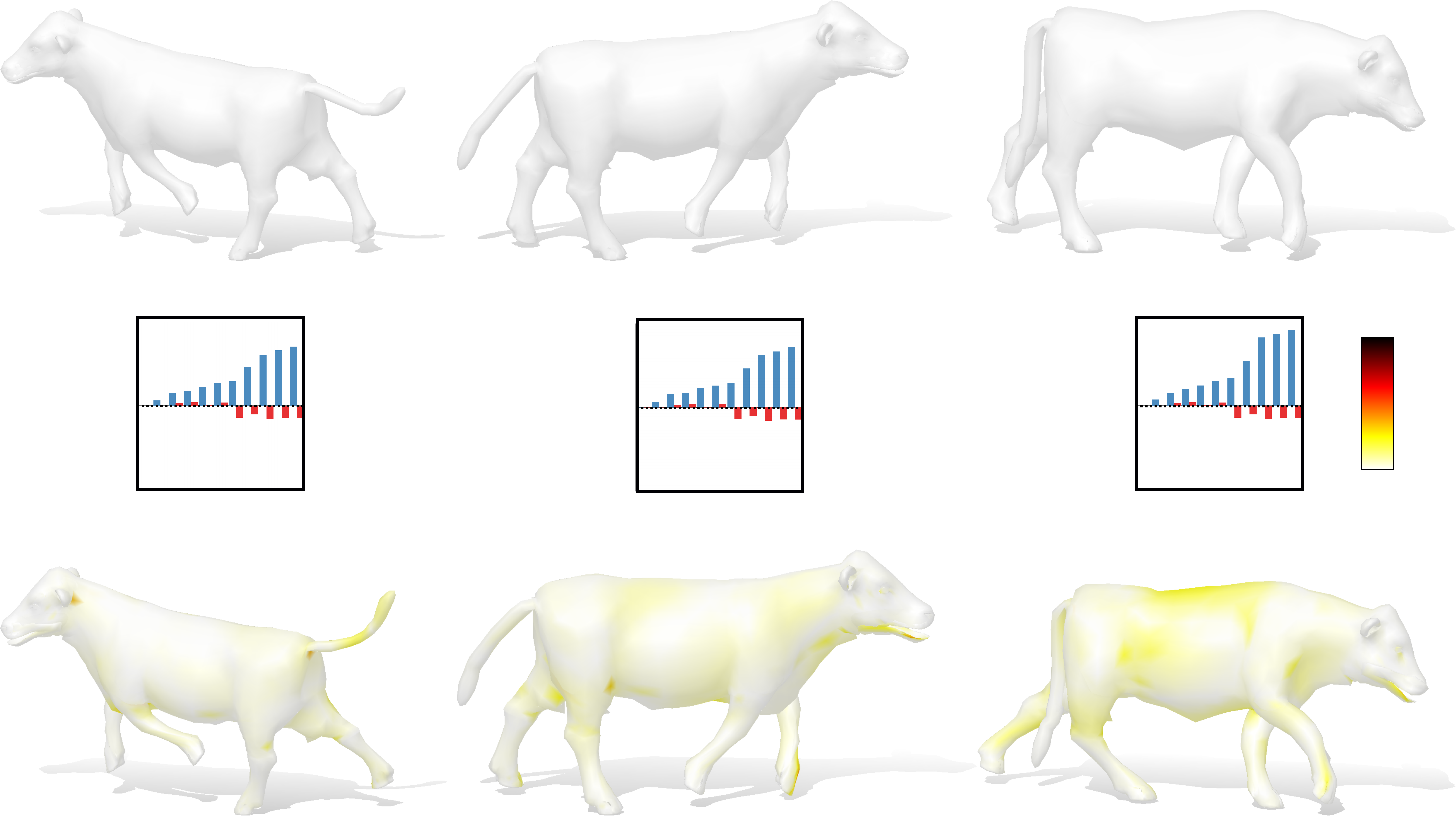}
    \put(10,56){\footnotesize{\darkgreen{cow}}}     
    \put(43,56){\footnotesize{\darkgreen{cow}}}
    \put(78,56){\footnotesize{\darkgreen{cow}}}
    \put(78,-2.5){\footnotesize{\darkred{big cat}}}
    \put(43,-2.5){\footnotesize{\darkred{big cat}}}
    \put(10,-2.5){\footnotesize{\darkred{big cat}}}     
    \put(76.2,28){\tiny $0$}   
    \put(72.3,22){\tiny $-28$}   
    \put(72.3,33){\tiny $+28$}  
    \put(41.5,28){\tiny $0$}   
    \put(37.6,22){\tiny $-28$}   
    \put(37.6,33){\tiny $+28$}  
    \put(7.5,28){\tiny $0$}   
    \put(3.6,22){\tiny $-28$}   
    \put(3.6,33){\tiny $+28$}  
    \put(97,24){\tiny $0$}   
    \put(97,32){\tiny $0.2$}  
   \end{overpic}
   \vspace{0.12cm}

    \begin{overpic}[trim=0cm 0cm 0cm 0cm,clip,width=\linewidth]{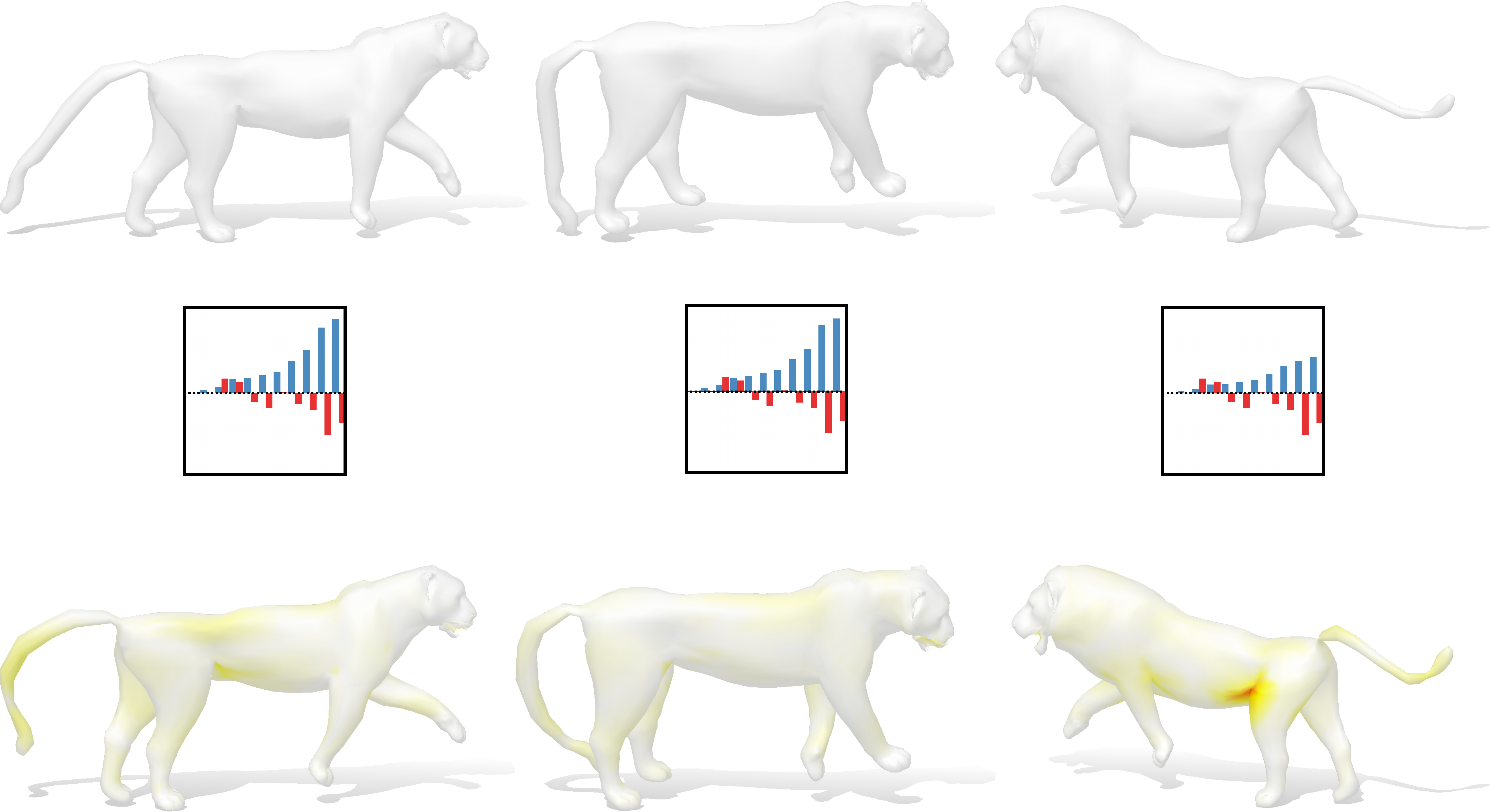}
    \put(0,57){\color{black}\line(1,0){100}}
    \put(10,54){\footnotesize{\darkgreen{big cat}}}     
    \put(43,54){\footnotesize{\darkgreen{big cat}}}
    \put(78,54){\footnotesize{\darkgreen{big cat}}}
    \put(78,-2.5){\footnotesize{\darkred{hippo}}}
    \put(43,-2.5){\footnotesize{\darkred{horse}}}
    \put(10,-2.5){\footnotesize{\darkred{horse}}}     
    \put(76.2,28){\tiny $0$}   
    \put(72.3,22){\tiny $-44$}   
    \put(72.3,33){\tiny $+44$}  
    \put(44.5,28){\tiny $0$}   
    \put(40.6,22){\tiny $-44$}   
    \put(40.6,33){\tiny $+44$}  
    \put(10.5,28){\tiny $0$}   
    \put(6.6,22){\tiny $-44$}   
    \put(6.6,33){\tiny $+44$}  
   \end{overpic}
\caption{\label{fig:SMALqualitative} Examples of universal adversarial attacks on PointNet from 2 classes of the SMAL dataset (top: cows, bottom: big cats). The heatmap encodes curvature distortion, growing from white to dark red. Even if the original shapes are not isometric, as can be noted also from their spectra (blue bars), a universal spectral perturbation $\rho$ (red bars, scaled by a factor $10^{3}$) leads to misclassification.
}
\end{figure}

\begin{figure*}[t]
\centering
     \begin{overpic}[trim=0cm 0cm 0cm 0cm,clip,width=\linewidth]{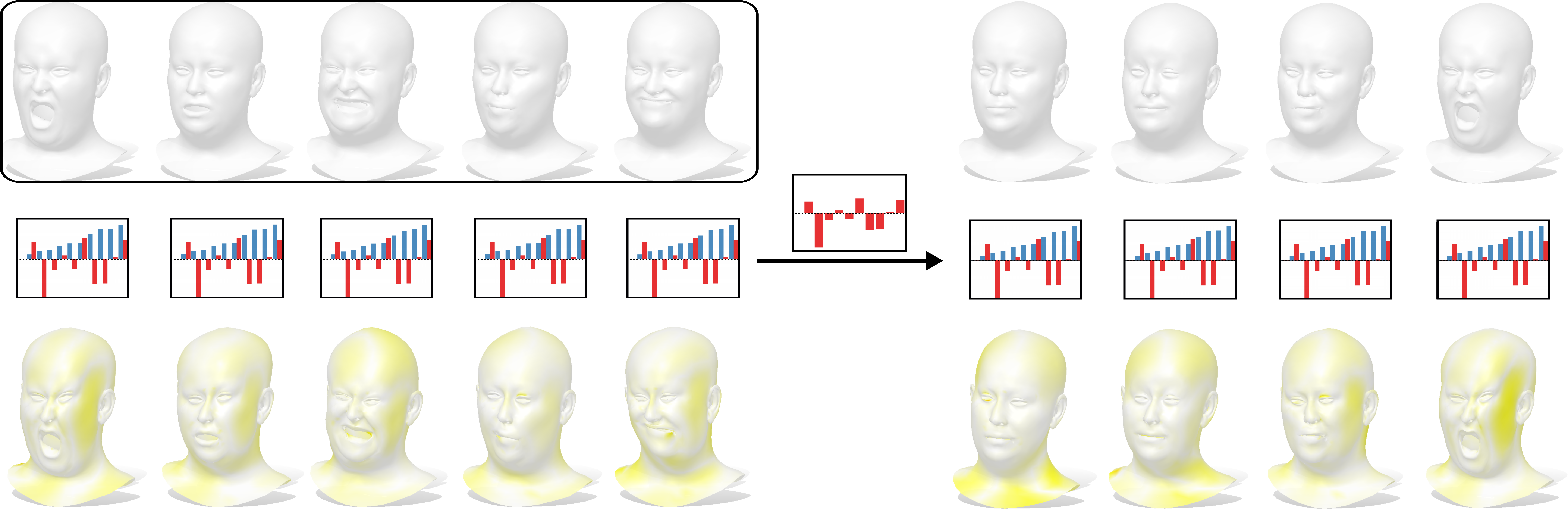}
     
     \put(15,33){\footnotesize{ Universal attack on \darkgreen{ID 10}}}

     \put(72,33){\footnotesize{ Generalization on \darkgreen{ID 10}}}
     
     \put(3,-1.5){\footnotesize{\darkred{ ID 2}}}
     \put(13,-1.5){\footnotesize{\darkred{ ID 2}}}
     \put(22.5,-1.5){\footnotesize{\darkred{ ID 3}}}
     \put(32.5,-1.5){\footnotesize{\darkred{ ID 3}}}
     \put(41,-1.5){\footnotesize{\darkred{ ID 2}}}
     
     \put(63,-1.5){\footnotesize{\darkred{ ID 3}}}
     \put(73,-1.5){\footnotesize{\darkred{ ID 3}}}
     \put(83,-1.5){\footnotesize{\darkred{ ID 2}}}
     \put(93,-1.5){\footnotesize{\darkred{ ID 3}}}
     
     \put(54,22){\footnotesize{$\rho$}}
    \put(58,18.6){\tiny $0$}   
    \put(58,16.5){\tiny $-0.01$}   
    \put(58,20.5){\tiny $+0.01$}
    
    \put(0,16.2){\tiny $0$}   
    \put(-1.3,18){\tiny $720$} 
     \end{overpic}
\caption{\label{fig:COMAqualitativeALL} On the left, examples of a universal adversarial attack on 15 shapes (only 5 shown) classified with PointNet. The resulting universal spectral perturbation (middle) is used to generalize the attack to 4 new shapes of the same subject (right).}
\end{figure*}
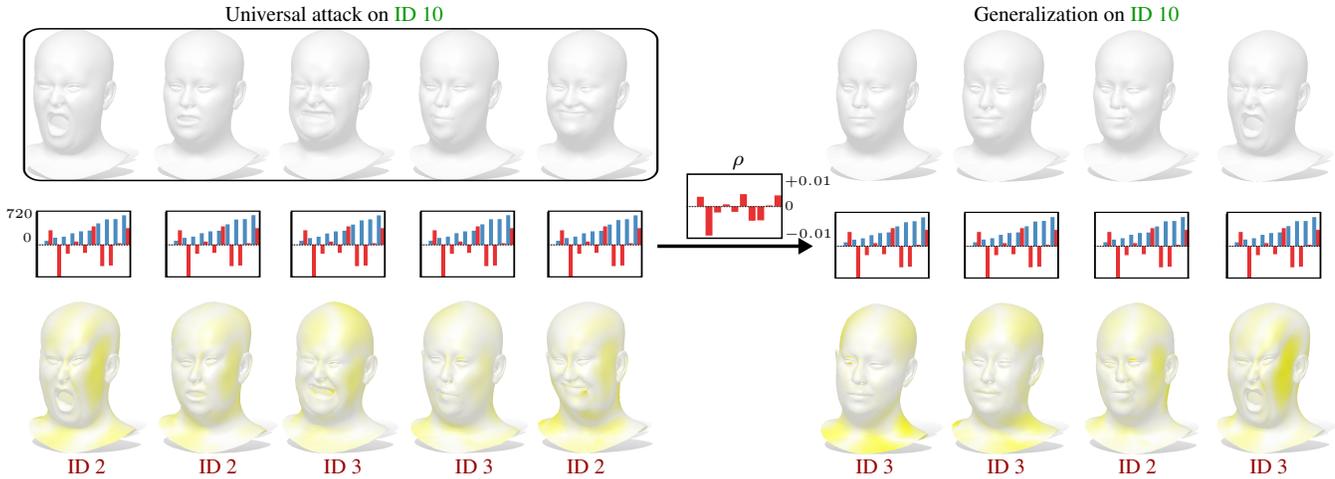

%
We optimize problem~\eqref{eq:general_formula3} over 15 random shapes of the same class from the test set, obtaining a set of spatial coefficients $\alpha_i$ for each shape $X_i$, and a universal perturbation $\rho$ common to all the shapes. We use the same optimization parameters for all the datasets and classifiers, with ${c=5e-2}$.
In Table~\ref{tab:comparison} we evaluate the quality of our adversarial attacks in terms of success rate and deformation strength. As we can see, the strength of the attack is inversely proportional to the success rate, further confirming the conclusions drawn in our sensitivity analysis. 
Table~\ref{tab:comparison} also includes results obtained with our method {\em without} optimizing for the universal perturbation $\rho$ in Eq.~\eqref{eq:general_formula3}, i.e., we optimize for the coefficients $\alpha_i$ independently for each shape. As expected, this leads to a slight increase of the success rate and a less noticeable deformation.
%
%
For completeness, we also report results from the state-of-the-art method~\cite{Mariani20}, which uses a geometric regularizer to bound the distortion.
Several qualitative examples are shown in Figures \ref{fig:COMAqualitative}, \ref{fig:SMALqualitative} and \ref{fig:COMAqualitativeALL}.



\begin{figure}[t]
\vspace{0.1cm}

\centering
 \begin{overpic}[trim=0cm 20cm 0cm 0cm,clip,width=\linewidth]{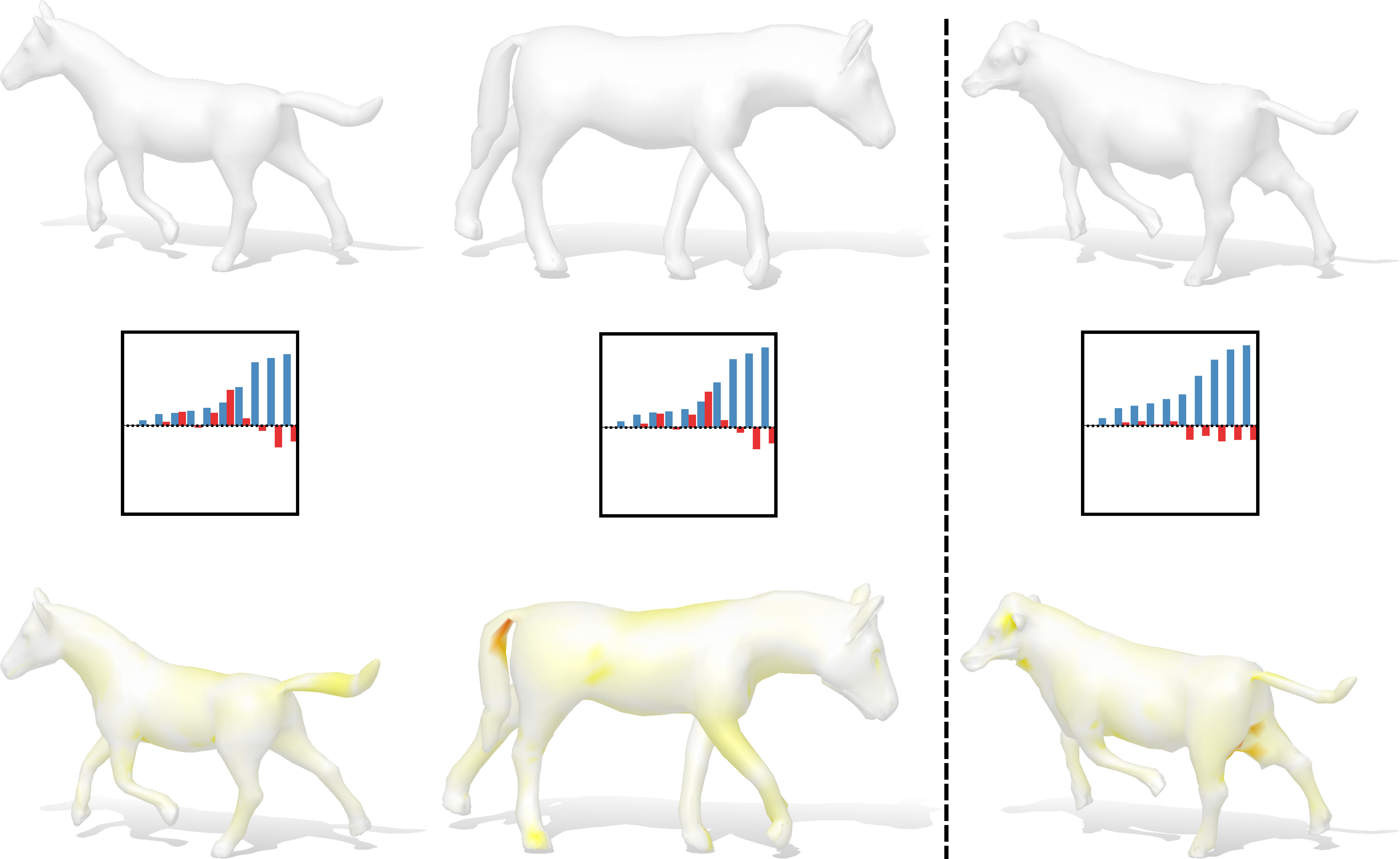}
 \end{overpic}
   \begin{overpic}[trim=0cm 0cm 0cm 35cm,clip,width=\linewidth]{./figures/SMALgeneral_sep.pdf}
    \put(9,58){\footnotesize{\textbf{attack}}}
    \put(39,58){\footnotesize{\textbf{generalization}}}
    \put(74.4,58){\footnotesize{\textbf{generalization}}}
    \put(43,37){\footnotesize{\darkgreen{horse}}}
    \put(10,37){\footnotesize{\darkgreen{horse}}}     
    \put(78,-1){\footnotesize{\darkred{horse}}}
    \put(43,-1){\footnotesize{\darkred{cow}}}
    \put(10,-1){\footnotesize{\darkred{cow}}}     
    \put(78,37){\footnotesize{\darkgreen{cow}}}
    \put(75.2,27){\tiny $0$}   
    \put(71.2,22){\tiny $-28$}   
    \put(71.3,33){\tiny $+28$}  
    \put(40.5,27){\tiny $0$}   
    \put(36.5,22){\tiny $-32$}   
    \put(36.6,33){\tiny $+32$}  
    \put(6.5,27){\tiny $0$}   
    \put(2.5,22){\tiny $-32$}   
    \put(2.6,33){\tiny $+32$}  
   \end{overpic}
\caption{\label{fig:SMALgeneralization} Generalization results on two different classes from SMAL. Left: the spectral perturbation $\rho$ estimated for a set of 15 horses (1 shown in the first column) is applied to an unseen shape of the same class (second column) as described in the text. The resulting deformed shape (second column, bottom shape) is incorrectly classified. Right: another example from a different class. In the bar plots $\rho$ (red) is scaled by a factor $10^{3}$.}
\end{figure}
%


%
\subsection{Generalization} 
%
As described in Section~\ref{sec:alg}, once we optimize for a universal perturbation $\rho$ on a set of shapes, this can  be used to transfer the deformation to a new shape.
We do this by smoothly deforming the new shape as to make its spectrum match the target perturbation $\rho$. 
\begin{wrapfigure}[6]{r}{0.4\linewidth}
\vspace{-0.7cm}
\begin{center}
\hspace{-0.8cm}
\input{./figures/generalization_loss.tikz}
\end{center}
\end{wrapfigure}
In the inset figure, we show 
 that minimizing only the spectral term in Eq.~\eqref{eq:general_formula3} induces by itself a minimization of the adversarial penalty, bringing in turn the classifier to a wrong prediction. This suggests that the spectral energy is a strong prior for finding adversarial perturbations.

We performed a quantitative evaluation of the generalization capability on the CoMA dataset, obtaining successful adversarial attacks on unseen shapes on 80.8\% of the cases for the PointNet classifier, and 49.2\% with ChebyNet.
Finally, in Figures~\ref{fig:SMALgeneralization} and~\ref{fig:COMAqualitativeALL} we show some qualitative examples on both the CoMA and SMAL datasets, showing how we are able to produce a similar deformation on new samples without requiring any correspondence.
%

%

%
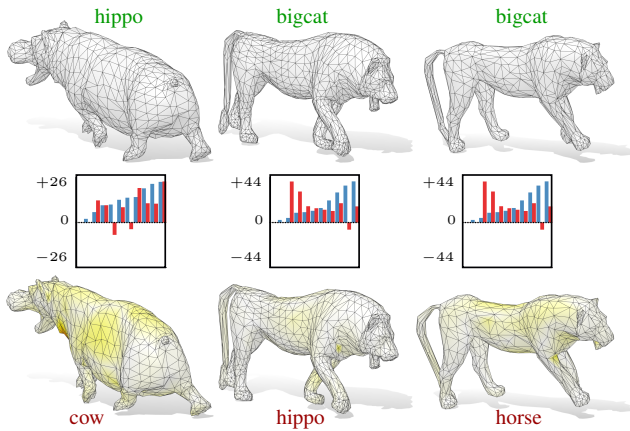
\begin{figure}[b]
\centering
     \begin{overpic}[trim=0cm 0cm 0cm 0cm,clip,width=\linewidth]{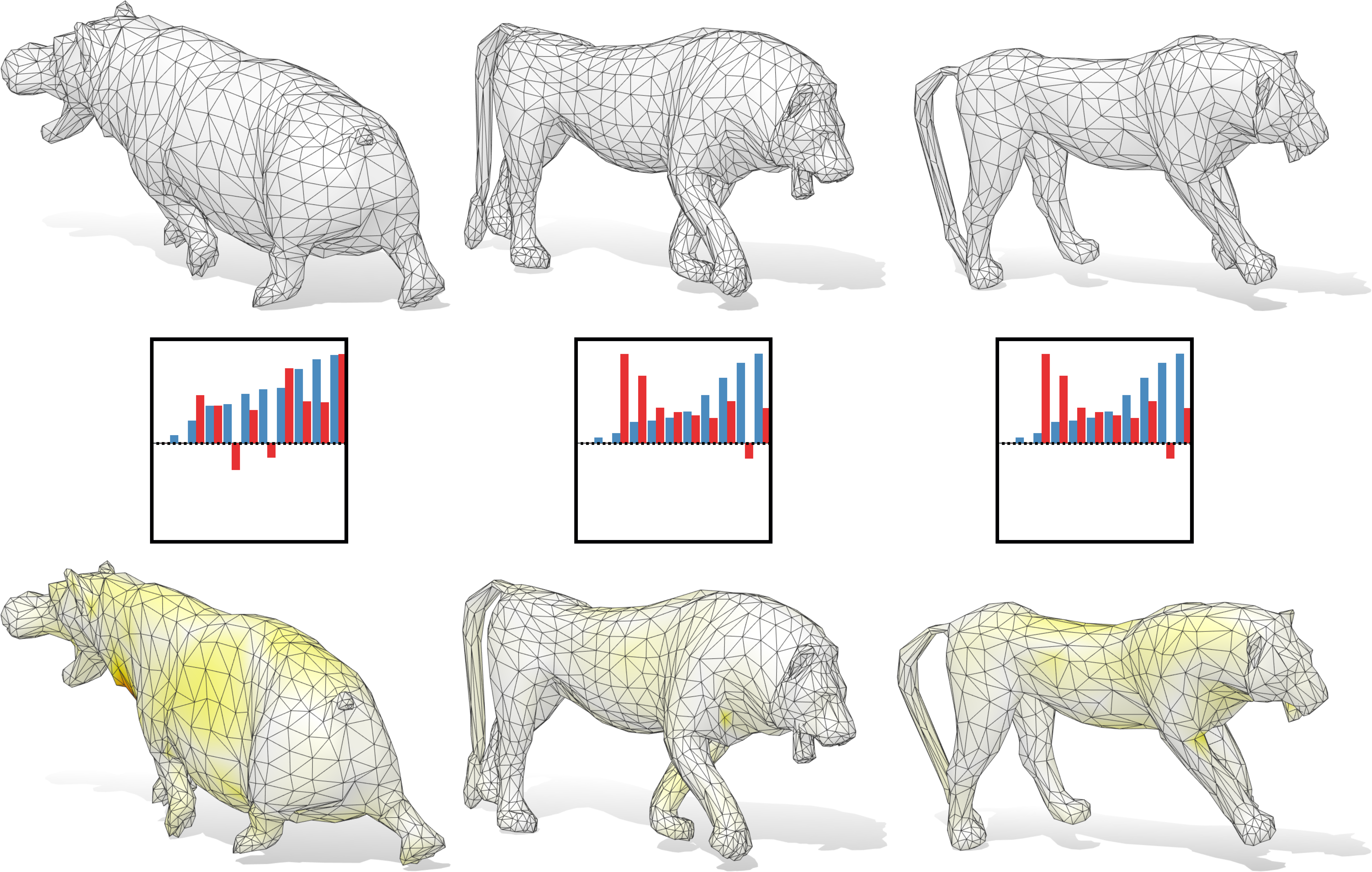}
    \put(14,63){\footnotesize{\darkgreen{hippo}}}     
    \put(43,63){\footnotesize{\darkgreen{bigcat}}}
    \put(78,63){\footnotesize{\darkgreen{bigcat}}}
    \put(78,-1){\footnotesize{\darkred{horse}}}
    \put(43,-1){\footnotesize{\darkred{hippo}}}
    \put(10,-1){\footnotesize{\darkred{cow}}}     
    \put(70.2,31){\tiny $0$}   
    \put(66.2,25){\tiny $-44$}   
    \put(66.3,37){\tiny $+44$}  
    \put(39.5,31){\tiny $0$}   
    \put(35.5,25){\tiny $-44$}   
    \put(35.6,37){\tiny $+44$}  
    \put(8.5,31){\tiny $0$}   
    \put(4.5,25){\tiny $-26$}   
    \put(4.6,37){\tiny $+26$}  
   \end{overpic}
\caption{\label{fig:remeshedSMAL} Results on remeshed shapes from SMAL. The first column shows the result of the universal attack on the hippo category, while the remaining two are on the bigcat category. Both are conducted on a set of 15 shapes. }
\end{figure}

\subsection{Meshes and point clouds}
One of the main advantages of working in a spectral domain is that it is agnostic to the surface representation, as long as a good approximation of the Laplacian operator can be computed on it. This property allows us to handle seamlessly shapes with different tessellation, resolution, and even surfaces represented by unorganized point clouds. In fact, the only limitation is posed by the classifier under attack, which might be representation-specific.

\vspace{1ex}\noindent\textbf{Triangle meshes.}
To prove the robustness of our attack to different tessellations, we  independently remeshed each shape of the SMAL dataset to random number of vertices within $30\%$ to $50\%$ of the original ones. We then used these shapes to perform a universal adversarial attack on the PointNet classifier. Not surprisingly, we noted an increase of performance for the adversarial attack, obtaining up to $94\%$ of success rate with an average curvature distortion of $0.80$. This improvement is explainable by the reduced number of points given as input to PointNet for classification, making the adversarial attack easier to perform. Qualitative examples are shown in Figure~\ref{fig:remeshedSMAL}.

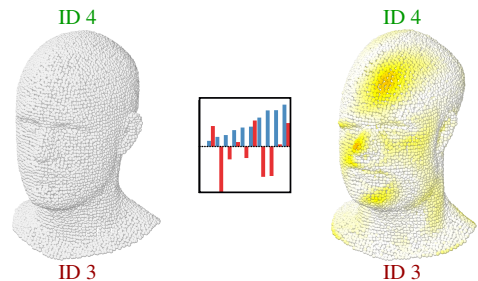
\begin{figure}[b]
\centering
     \begin{overpic}[trim=0cm 0cm 0cm 0cm,clip,width=0.74\linewidth]{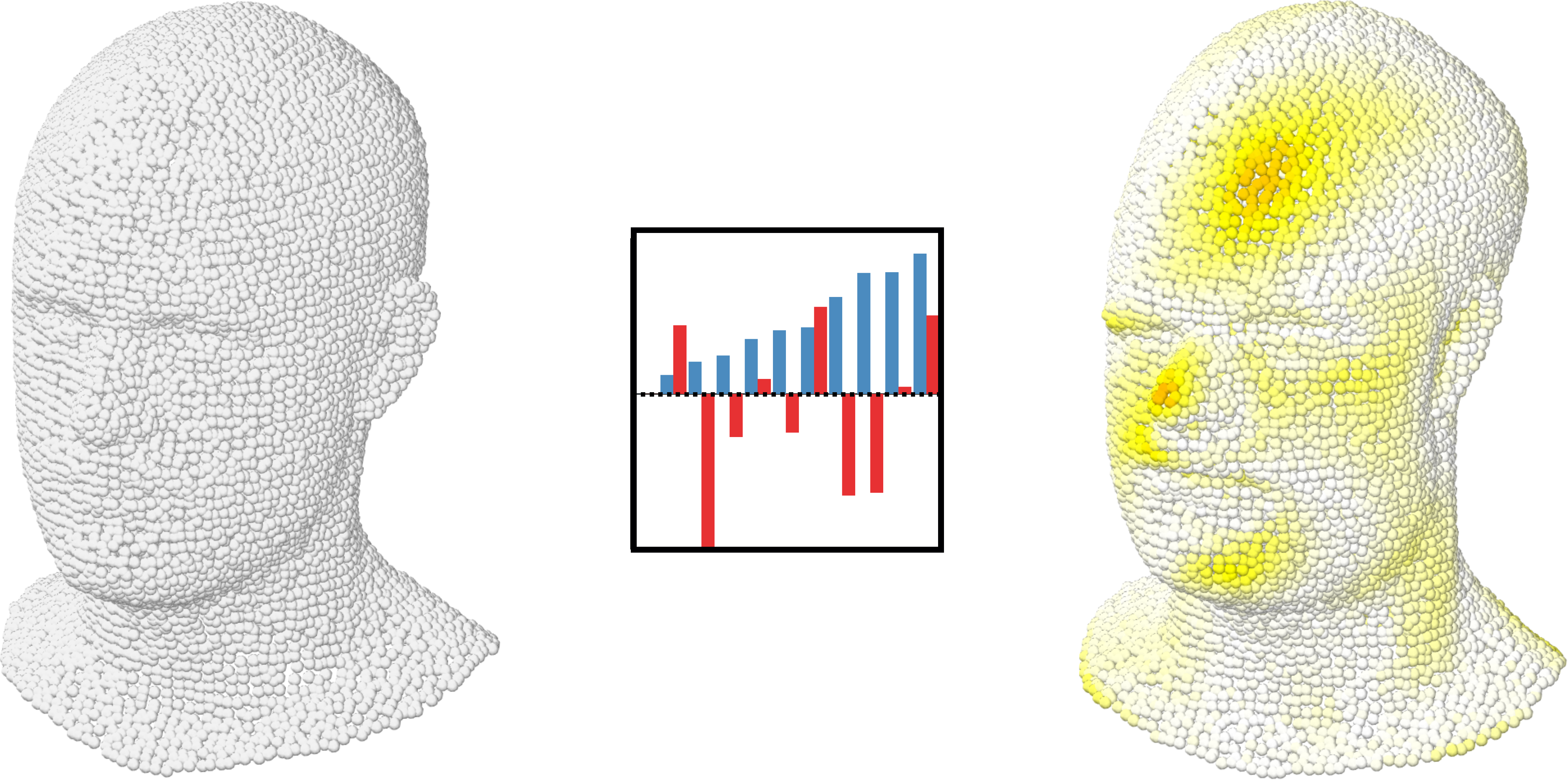}
     \put(10,51){\footnotesize{\darkgreen{ID 4}}}
     \put(10,-4){\footnotesize{\darkred{ID 3}}}
     \put(80,51){\footnotesize{\darkgreen{ID 4}}}
     \put(80,-4){\footnotesize{\darkred{ID 3}}}
  \end{overpic}
\caption{\label{fig:pcCOMA} Example of generalization to a point cloud in CoMA using PointNet classifier.}
\end{figure}

\vspace{1ex}\noindent\textbf{Point clouds.}
In Figure~\ref{fig:pcCOMA} we show a qualitative example of generalization to point clouds. We optimized for a universal perturbation to the PointNet classifier, using 15 meshes from the CoMA test set. We then applied our generalization procedure to a point cloud derived from a new pose of the same subject. To estimate a Laplace operator for the point cloud, we used the method described in~\cite{clarenz2004finite}.

%% file: sections/conclusion.tex
\section{Conclusion}\label{sec:conclusion}
\vspace{-0.6ex}
We introduced a method to compute universal adversarial perturbations on 3D geometric data. The key idea lies in the adoption of the Laplacian spectrum as an intermediate shared domain for multiple shapes, where perturbations can be computed and then resynthesized into the geometry via shape-from-spectrum recovery. Operating with eigenvalues endows our attacks with robustness to deformation, sampling, and shape representation, leading in turn to generalization outside of the optimization set.
Currently, the main \textbf{limitation} of this approach is its limited applicability to shapes belonging to very different classes; for example, we were not able to find successful universal perturbations for faces {\em and} animals simultaneously. This is due to the fact that different classes may have very different spectra; looking for an alternative, perhaps learned, representation might be a potential solution to explore in the future.


%% file: sections/architectures.tex
\section{Architecture of the classifiers}



Two types of networks were used in the tests, depending on the input. For meshes the implemented architecture is similar to the state-of-the-art encoder used by the CoMA autoencoder \cite{COMA}. The structure of the network is shown in Table~\ref{tab:cheb-architecture}. It consists of 4 layers of ReLU-activated fast Chebyshev filters \cite{defferrard2016convolutional} with size $K = 6$, interleaved by mesh decimation via iterative edge collapse \cite{garland1997surface}, and a final dense layer. We refer to this classifier simply as \textbf{ChebyNet}.

\begin{table}[h]
	\centering
	\begin{tabular}{lcc}
		Layer         & Input Size & Output Size\\
		\hline
		Convolution   & 3889 $\times$ 3   & 3889 $\times$ 128\\
		Down-sampling & 3889 $\times$ 128 & 1945 $\times$ 128\\
		Convolution   & 1945 $\times$ 128 & 1945 $\times$ 128\\
		Down-sampling & 1945 $\times$ 128 & 973  $\times$ 128\\
		Convolution   & 973  $\times$ 128 & 973  $\times$ 64\\
		Down-sampling & 973  $\times$ 64  & 487  $\times$ 64\\
		Convolution   & 487  $\times$ 64  & 487  $\times$ 64\\
		Fully Connected & 31168 & $|\mathcal{C}|$ 
	\end{tabular}
\caption{ChebyNet classifier architecture in detail for the SMAL dataset. $n=3889$ is the number of vertices for the input meshes, and $|\mathcal{C}|$ is the number of classes.}
\label{tab:cheb-architecture}
\end{table}

For point clouds we used the \textbf{PointNet} classifier \cite{qi2017pointnet}, composed by 4 layers of point convolution followed by batchnorm with ReLU, with layer output sizes $32 \veryshortarrow 128 \veryshortarrow 256 \veryshortarrow 512$. A maxpool operation is used to output a 512-dimensional vector, which is then reduced with a ReLU-activated fully connected network to dimensions: $512 \veryshortarrow 256 \veryshortarrow 128 \veryshortarrow 64 \veryshortarrow |\mathcal{C}|$, where $|\mathcal{C}|$ is the number of classes. 
Both ChebyNet and PointNet are trained to classify the subject identity for shapes in CoMA  dataset \cite{COMA}, and the animal species for shapes in SMAL dataset \cite{zuffi20173d}. The accuracy achieved in each case is reported in Table \ref{tab:accuracy}.


\begin{table}[]

	\centering
	\begin{tabular}{l|cc}
	    \hline
		\textbf{SMAL}     & ChebyNet & PointNet \\
		\hline
		train    & 100\% & 98.1\%  \\
		test     & 100\% & 94.2\%  \\
		remeshed & -     & 88.3\% \\
		\hline
		\hline
		\textbf{CoMA}  & ChebyNet & PointNet \\
		\hline
		train &  99.6\%   & 99.0\%  \\
		test  &  99.0\%   & 99.2\%  \\	
		\hline
	\end{tabular}
	\caption{Accuracy of the four considered classifiers in terms of fraction of correct predictions. For the SMAL dataset, PointNet was evaluated also on remeshed shapes from the test set, with a random number of vertices within 30\% to 50\% of the original ones.}
	\label{tab:accuracy}
\end{table}

%% file: sections/additional_results.tex
\section{Additional results}\label{sec:results}

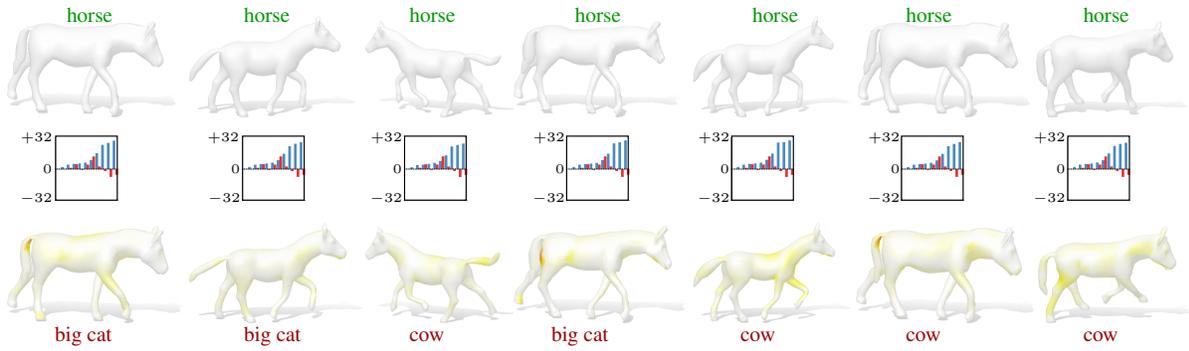
\begin{figure*}[t]
\centering
    \begin{overpic}[trim=0cm 0cm 0cm 0cm,clip,width=0.9\linewidth]{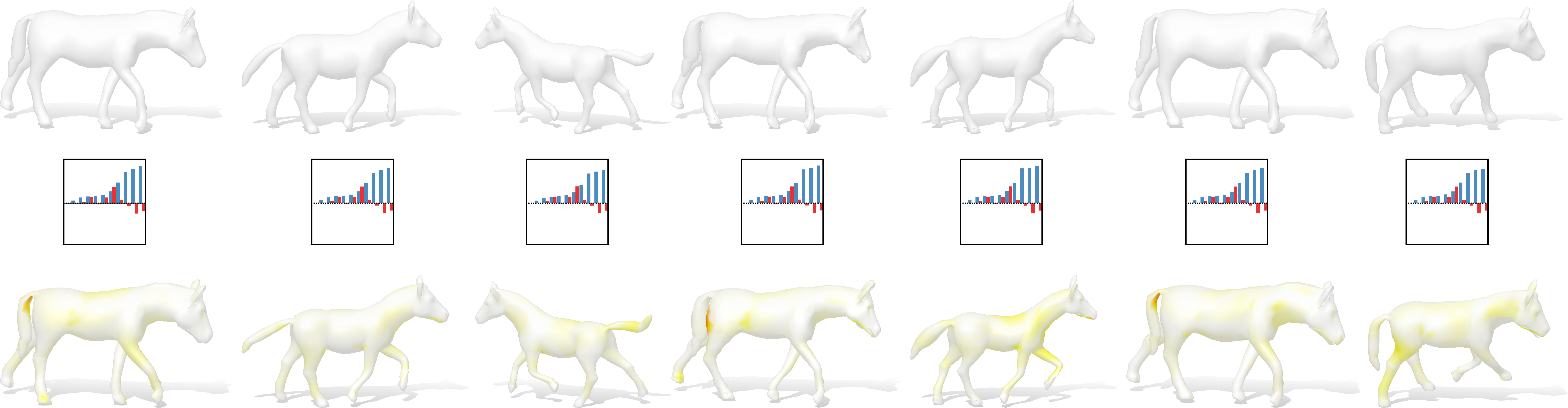}
      \put(5,25.5){\footnotesize{\darkgreen{horse}}}       \put(20,25.5){\footnotesize{\darkgreen{horse}}}       \put(34,25.5){\footnotesize{\darkgreen{horse}}}       \put(48,25.5){\footnotesize{\darkgreen{horse}}} 
      
      \put(63,25.5){\footnotesize{\darkgreen{horse}}}       \put(76,25.5){\footnotesize{\darkgreen{horse}}}       \put(91,25.5){\footnotesize{\darkgreen{horse}}}     
    \put(4,-1.5){\footnotesize{\darkred{big cat}}}       \put(20,-1.5){\footnotesize{\darkred{big cat}}}       \put(34,-1.5){\footnotesize{\darkred{cow}}}       \put(46,-1.5){\footnotesize{\darkred{big cat}}} 
      
      \put(62,-1.5){\footnotesize{\darkred{cow}}}       \put(76,-1.5){\footnotesize{\darkred{cow}}}       \put(91,-1.5){\footnotesize{\darkred{cow}}}     
    \put(3,12.75){\tiny $0$} \put(1.05,10.3){\tiny $-32$}   
    \put(1.1,15.4){\tiny $+32$}  
    
    \put(18.9,12.75){\tiny $0$} \put(16.95,10.3){\tiny $-32$}   
    \put(16.9,15.4){\tiny $+32$}  
    
    \put(32.6,12.75){\tiny $0$} \put(30.65,10.3){\tiny $-32$}   
    \put(30.7,15.4){\tiny $+32$}  
     
    \put(46.35,12.75){\tiny $0$} \put(44.4,10.3){\tiny $-32$}   
    \put(44.45,15.4){\tiny $+32$}  
    
    \put(60.3,12.75){\tiny $0$} \put(58.35,10.3){\tiny $-32$}   
    \put(58.4,15.4){\tiny $+32$}  
    
    \put(74.7,12.75){\tiny $0$} \put(72.75,10.3){\tiny $-32$}   
    \put(72.8,15.4){\tiny $+32$}  
    
    \put(88.7,12.75){\tiny $0$} \put(86.75,10.3){\tiny $-32$}   
    \put(86.8,15.4){\tiny $+32$}  
    
    \end{overpic}
\caption{\label{fig:HORSEuniversal} Example of universal adversarial attacks on PointNet over 7 shapes from the horse class of SMAL. The heatmap encodes curvature distortion, growing from white to dark red. Even if the original shapes are not isometric, as can be noted also from their spectra (blue bars), a universal spectral perturbation $\rho$ (red bars, scaled by a factor $10^{3}$) leads to misclassification.}
\end{figure*}

In Fig.~\ref{fig:HORSEuniversal} we show additional qualitative examples of universal attacks that due to lack of space were not included in the main manuscript.

\vspace{1ex}\noindent
\textbf{Number of eigenvalues.} 
We performed an analysis of the generalization capability of our method to previously unseen shapes at varying number of eigenvalues $k$. For each class we considered 15 shapes on which we performed the universal attack. We then transfer the deformation to 10 new shapes of the same class. Results on the \textbf{CoMA} dataset are reported in Table \ref{tab:k}.
Considering a larger number of eigenvalues $k$ leads to an increase of success rate for the generalization. After $k=60$, the performance decreases due to the difficulty to transfer the spectral deformation $\rho$, as measured by the \textit{alignment error} $\epsilon_i = \| \sigma(X_i)(1+\rho) - \sigma(X_i + {\Phi}_i{\alpha}_{i}) \|$, where $X_i$ is the original shape geometry and $\alpha_i$ are the perturbation coefficients. Since from the perturbed eigenvalues we synthesize novel shapes (the adversarial examples), we cannot compute a geometric error because a ground-truth 3D reconstruction does not exist.  However, we can measure the alignment between the spectrum of the synthesized shapes and the target perturbed eigenvalues.

\begin{figure}[t]
\centering
    \begin{overpic}[trim=0cm 0cm 0cm 0cm,clip,width=0.85\linewidth]{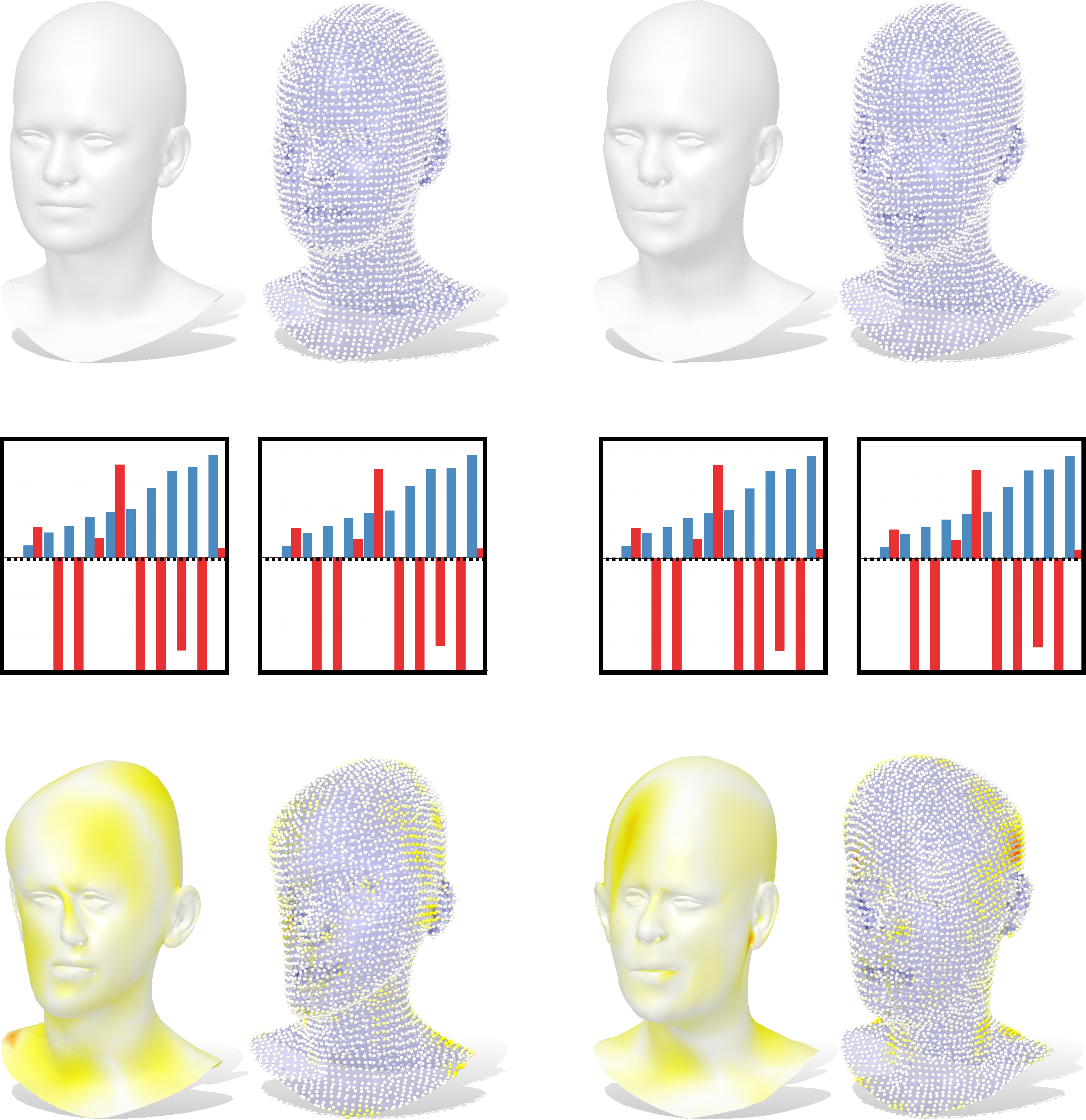}
       
    
    \put(16.5,99){\footnotesize{\darkgreen{ID 2}}} \put(68,99){\footnotesize{\darkgreen{ID 2}}}    
    \put(17.5,-3){\footnotesize{\darkred{ID 6}}}   \put(68.5,-3){\footnotesize{\darkred{ID 6}}}   
    \put(-2,49.2){\tiny $0$} \put(-7.75,39.5){\tiny $-670$}   
    \put(-7.8,60){\tiny $+670$}  
    
    \put(51.2,49.2){\tiny $0$} \put(45.45,39.5){\tiny $-670$}   
    \put(45.4,60){\tiny $+670$}  
    \end{overpic}
\caption{\label{fig:COMApcs} Examples of generalization to point clouds. The spectral perturbation $\rho$ (red bars, scaled by a factor $10^3$) was obtained on a set of 15 meshes (not shown). The deformation was then transferred to 2 unseen shapes discretized both as meshes (white on the left) and as point clouds (light blue on the right). The deformed shapes are shown in the last row. As we can see, for each shape the deformations induced by $\rho$ are approximately the same regardless of the discretization. Note that here we intentionally enhanced the strength of the deformation (by increasing the weight of the adversarial loss) to better appreciate the similarity between the mesh and point cloud cases.}
\end{figure}
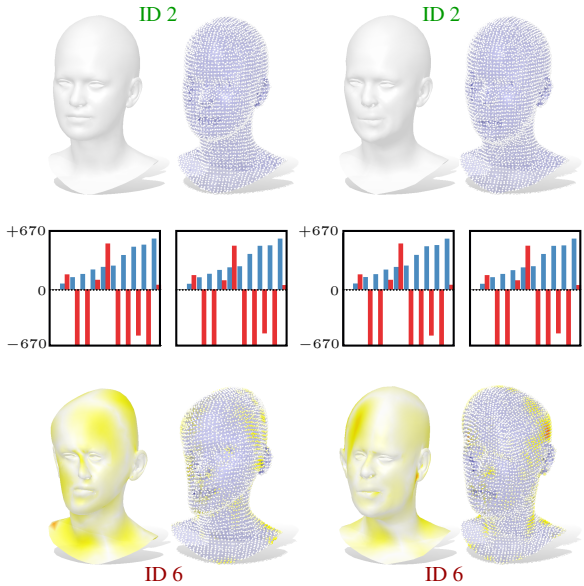

\vspace{1ex}\noindent
\textbf{Point clouds.}
In Fig.~\ref{fig:COMApcs} we show another example of generalization to point clouds; we compare the resulting deformation with the same perturbation applied to the corresponding mesh. To better appreciate the similarity between the two we exaggerated the perturbation of the universal attack by increasing the weight of the adversarial loss $c$.

\begin{table}
	\centering
    \begin{tabular}{r|cc}
	    \hline
		 $k$    & success rate  & alignment error \\ 
		\hline
		10 & 12\% & 2.65e-4\\
		20 & 56\% & 1.33e-4\\
		30 & 61\% & 1.96e-4\\ 
		40 & 80\% & 2.72e-4\\
		60 & 78\% & 3.06e-4\\
		80 & 49\% & 5.84e-4\\
		100& 17\% & 7.01e-4\\
		\hline
	\end{tabular}
	
	\caption{Dependence of the generalization capability of our method on the number of used eigenvalues $k$. The \emph{alignment error} is the absolute error between the target eigenvalues computed with $\rho$, and the eigenvalues of the deformed shapes; the \emph{success rate} is the percentage of attacks that induce misclassification.}
	\label{tab:k}
\end{table}

